\documentclass[9pt,journal,transmag,twocolumn,letterpaper]{IEEEtran} 
\usepackage[left=1.60cm,right=1.60cm,top=0.95cm,bottom=2.54cm]{geometry}

\usepackage{amsthm}

\usepackage{microtype} 
\usepackage{siunitx} 

\usepackage{tabularx,booktabs}
\usepackage[T1]{fontenc}
\usepackage[ruled,vlined]{algorithm2e}
\usepackage{algcompatible}
\usepackage{enumitem}
\usepackage{algorithmicx}
\usepackage{cite}
\usepackage[colorlinks,urlcolor=blue,linkcolor=blue,citecolor=blue]{hyperref}
\usepackage{graphicx}
\usepackage{booktabs, makecell}
\usepackage{threeparttable}

\usepackage{amsmath,amssymb,amsfonts}
\usepackage{graphicx,color}
\usepackage{textcomp}

\usepackage{booktabs,arydshln}
\usepackage{tabularray}
\usepackage{multirow}
\usepackage{mathtools}
\usepackage{float}

\newcommand{\myremark}[1]{\textbf{\textit{Remark: }}#1}

\setcellgapes{3pt}

\newcolumntype{C}[1]{>{\centering\arraybackslash}p{#1}}
\newcolumntype{L}{>{\raggedright\arraybackslash}X}
\usepackage{array}
\usepackage[utf8]{inputenc}

\def\BibTeX{{\rm B\kern-.05em{\sc i\kern-.025em b}\kern-.08em   
    T\kern-.1667em\lower.7ex\hbox{E}\kern-.125emX}}  
 
\author{
\IEEEauthorblockN{Orson~Mengara\textsuperscript{1 } } 
\IEEEauthorblockA{
    \textsuperscript{1} INRS-EMT, University of Québec, Montréal, QC, Canada.  \\
    \{\texttt{orson.mengara@inrs.ca}\}
}
}

\begin{document}

\markboth{Last update preprint, journal name -------, VOL.~.., NO.~..., month~2024
}{Orson   \MakeLowercase{\textit{et al.}}: Robust backdoor attack via diffusion models and bayesian approach } 

\title{ The last Dance : Robust backdoor attack via diffusion models and bayesian approach}

\maketitle

\begin{abstract}
Diffusion models are state-of-the-art deep learning generative models that are trained on the principle of learning forward and backward diffusion processes via the progressive addition of noise and denoising. In this paper, we aim to fool audio-based DNN models, such as those from the Hugging Face framework, primarily those that focus on audio, in particular transformer-based artificial intelligence models, which are powerful machine learning models that save time and achieve results faster and more efficiently. We demonstrate the feasibility of backdoor attacks (called `BacKBayDiffMod`) on audio transformers\cite{islam2023comprehensive}\cite{jain2022hugging} derived from Hugging Face, a popular framework in the world of artificial intelligence research. The backdoor attack developed in this paper is based on poisoning model training data uniquely by incorporating backdoor diffusion sampling and a Bayesian approach to the distribution of poisoned data.
\end{abstract}


\section{Introduction} 

\scalebox{2.9}{D}iffusion models \cite{sohl2015deep} have attracted particular attention recently in different fields or tasks, such as density estimation \cite{croitoru2023diffusion}, text \cite{li2022diffusion}, audio \cite{kong2020diffwave}, image synthesis \cite{ho2020denoising},\cite{dhariwal2021diffusion},\cite{song2020denoising}, text-to-image \cite{rombach2022high},\cite{zhai2023text},\cite{ramesh2022hierarchical}, text-to-audio conversion \cite{liu2023audioldm}, video  \cite{ho2022imagen,mei2023vidm}, audio generation \cite{kong2020diffwave} and financial data generation \cite{sattarov2023findiff}. Various downstream services begin to incorporate well-trained large diffusion models, like
Stable Diffusion \cite{rombach2022high}. 

In general, diffusion models (which are generative deep learning models that are trained on the principle of forward and backward diffusion processes via the progressive addition of noise and denoising) consider sample generation as a diffusion process modeled by stochastic differential equations (SDEs) \cite{song2020score}. However, diffusion models are known to suffer from a lack of convergence speed due to the need to generate samples from an approximation of the data via Markov chain Monte Carlo (MCMC) methods, thus requiring thousands of steps to converge to a sampling process. In the literature, various attempts have been made to solve this problem, such as the DDIM approach \cite{song2020denoising}, DDPM \cite{ho2020denoising}, SMLD \cite{song2019generative}, and Analytic-DPM \cite{bao2022analytic}, but these methods have not been effective. Because, these methods are often based on beliefs that are not entirely well-founded, such as the use of the generation process via reverse Brownian motion. This study is the first to assess the robustness of diffusion models to backdoor attacks on transformer-based deep pre-trained models (TB-DPTMs) \cite{zhang2021transformer},\cite{chen2021pre},\cite{han2023survey}, (transferred models). We survey the weaknesses and perils of diffusion models built on TB-DPTMs to study their vulnerability and danger to backdoor attacks \cite{chen2017targeted}. We introduce BacKBayDiffMod (Backdoor-based Bayesian Diffusion Model), a new attack approach that poisons the training process of a Bayesian diffusion model by constructing a backdoor Bayesian diffusion process at the test stage. At the inference step, the corrupted Bayesian diffusion model will behave normally for normal data inputs, while fraudulently producing a targeted output created by the malicious actor when it receives the trigger signal placed in the training data. Recently, diffusion models have been extended to automatic speech recognition tasks, as studied in \cite{malik2023preliminary},\cite{movellan1999diffusion},\cite{he2023voiceextender}. Backdoor attacks can also extend to other attack surfaces, such as speech control systems (such as Amazon's Alexa, Apple's Siri, Google Assistant, and Microsoft's Cortana, which rely on authentication or biometric identification) and home robotics scenarios \cite{nair2023environment}, \cite{chu2009environmental} such as the optimization of autonomous cars or autonomous robots capable of assisting humans in various tasks, which rely heavily on environmental sound datasets to train DNN models to make them operational and efficient for tasks such as classification. What's more, if a backdoor attack were to be launched on these infrastructures, the repercussions and damage caused would be very serious and enormous (this could lead, for example, to misinterpretation of sounds, posing a risk to road safety, or trigger false alarms, leading to dangerous reactions or even unnecessary allocation of resources in the case of domestic robots), both in terms of property and personal safety.

\vspace{1mm}

For example, with backdoor attacks \cite{yerlikaya2022data} (which often occur at training time when the model developer outsources model training to third parties), the attacker can insert a hidden behavior (called a backdoor) into the model so that it behaves normally when benign samples are used but makes erroneous decisions if a specific trigger is present in the test sample, such as an imperceptible pattern in the signal. These attacks may have negative effects on machine learning systems' dependability and integrity. These consist of improper prediction (detection or classification), illegal access, and system manipulation.
 
\vspace{1mm} 

In this paper, we present a new paradigm for creating a robust clean label backdoor attack (algorithm \ref{alg:poisoning_attack}), incorporating a Bayesian approach \cite{rios2023adversarial}, \cite{eriksson1994monte} (using a Fokker-Planck equation (algorithm \ref{alg:poisoning_attack_bayesian}) \cite{reich2021fokker},\cite{risken1996fokker}, for sampling via Yang-Mills theroy\footnote{\href{https://www.youtube.com/watch?v=2bngzq2ZZXY}{Kavli Hausdorff Center for Mathematics}} Simulator\footnote{\href{https://en.wikipedia.org/wiki/Stochastic_quantum_mechanics}{Stochastic quantum mechanics}} (algorithm \ref{alg:yms})\cite{kuipers2023quantum}. We also use a diffusion model (random Gaussian noise) \cite{chou2023villandiffusion},\cite{an2023elijah},\cite{chou2023backdoor}, and sampling approach (algorithm \ref{alg:improved_diffusion_sampling}), which implements a type of diffusion process (algorithm \ref{alg:improved_diffusion_sampling}) \cite{wang2024stronger}, i.e., a stochastic process that starts from a deterministic initial state and evolves according to a stochastic differential equation) to launch an audio backdoor attack on various automatic speech recognition audio models based on Hugging Face Transformers \cite{islam2023comprehensive} \cite{jain2022hugging}.

\section{Backdoor attack Machine Learning } 

In Machine Learning as a Service (MLaaS) applications, when the dataset, model, or platform are outsourced to unreliable third parties, backdoor attacks are increasingly serious and dangerous threats to machine learning systems. Most backdoor attacks, in general, insert hidden functions into neural networks by poisoning the training data; these are known as poisoning-based \cite{qiu2022survey}, \cite{li2022backdoor} backdoor attacks. The attacker applies a specific trigger pattern $T(\cdot)$ to partial samples in the dataset, $x^{\prime}=T(x)$, and gives a target label $y^{\prime}$ to them. The infected model $f_{\vartheta^{\prime}}$ would develop a hidden link between the trigger pattern and the target label after training on both the benign and poisoned samples. During the inference phase, the attacker might use the identical trigger sequence to activate the concealed backdoor. Consequently, the compromised model functions properly with benign inputs but reacts maliciously with certain triggers. Apart from being efficient, a typical backdoor attack also aims to be as stealthy as possible, meaning that it should not interfere with the system's regular operations and should not be easily seen by humans. Such an attack can be formulated as the following multi-objective optimization problem:

$$
\begin{aligned}
\underset{\vartheta^{\prime}}{\arg \min } \mathbb{E}_{\left(x^{\prime}, y^{\prime}\right) \in \mathcal{D}_p,(x, y) \in \mathcal{D}_b} & {\left[\mathcal{L}\left(f_{\vartheta^{\prime}}\left(x^{\prime}\right), y^{\prime}\right)\right.} \\
& \left.+\Lambda_1 \mathcal{L}\left(f_{\vartheta^{\prime}}(x), y\right)+\Lambda_2 D\left(x, x^{\prime}\right)\right],
\end{aligned}
$$

where $\mathcal{D}_p$ and $\mathcal{D}_b$ represent the subsets of poisoned and benign data. $D$ refers to the distance metric and $\Lambda_1$ and $\Lambda_2$ are weight parameters.

\subsection{Threat Model.} 

\textbf{Attack scenario.} In this paper, we start from a general scenario in which an audio system is deployed in any space, such as integrated speech control portals in smart homes \cite{Brilliant2023},\cite{Samsung2023}, and personal voice assistants on smart speakers \cite{Amazon2023},\cite{Google2023},\cite{Apple2023},\cite{zhai2021backdoor}. Third-party providers often create and implement the audio system. We examine an opponent that maliciously leverages the target audio system's backdoor vulnerability. The attacker may be a data or speech model editor, a system provider's employee, or even the supplier itself, with the potential to poison the data for backdoor insertion.

\vspace{1mm}

\textbf{Adversary objective.} In this paper, the adversary intends to implement a hidden backdoor in the target audio systems (e.g., speech control systems;  speech command recognition `SCR`, speech verification `SV`). By activating the backdoor in the system ASR, the adversary aims to inject malicious commands or impersonate legitimate users. At the same time, to keep the attack stealthy, the adversary seeks to minimize the audibility of the injected triggers and the impact on the normal operation of the target audio system.

\begin{figure}
\centering
\includegraphics[width=0.38\textwidth]{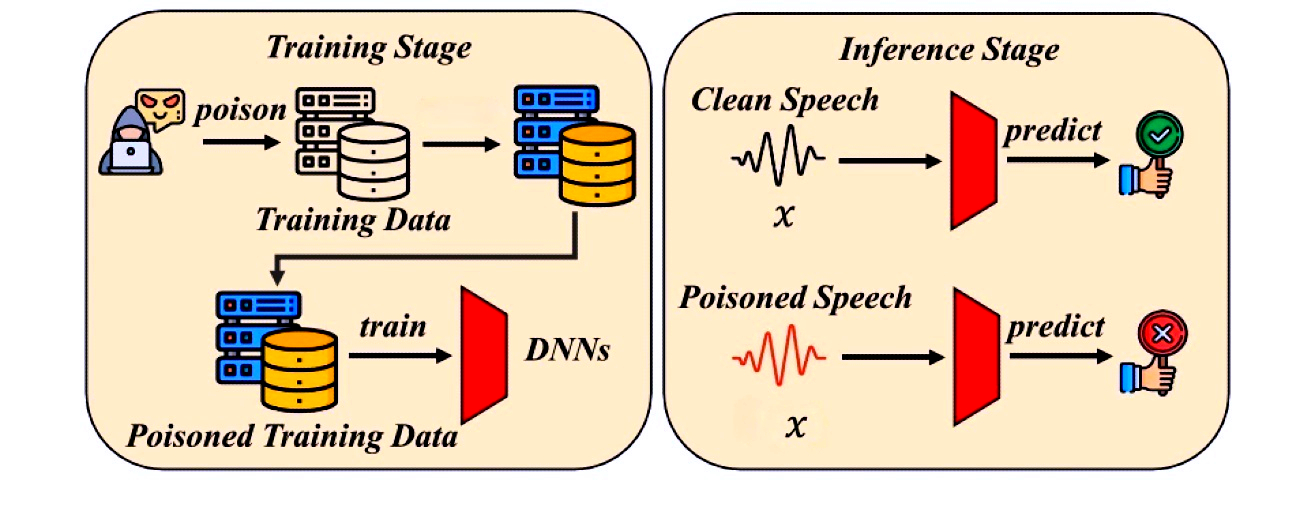}
\caption{Illustrates the execution process of a backdoor attack. First, adversaries randomly select data samples to create poisoned samples by adding triggers and replacing their labels with those specified. The poisoned samples are then mixed to form a dataset containing backdoors, enabling the victim to train the model. Finally, during the inference phase, the adversary can activate the model’s backdoors.}
\label{fig:backdoorexp}
\end{figure}

\vspace{1mm}

\textbf{Adversary capability.} We assume in this paper that the attacker (Figure \ref{fig:backdoorexp}) can only access the training dataset and undertake data poisoning, which is the most stringent parameter of backdoor attacks \cite{li2022backdoor}. The attacker has no prior knowledge of the target audio system's implementation details, such as preprocessing methods, network architectures, model parameters, learning methodologies, and so on.

\vspace{1mm}

\textbf{Model training.} The extracted acoustic features are fed to deep neural networks (DNNs) to infer the audio speech or speaker identity. Generally, for a dataset $\mathcal{D}=\{(x, y) \mid x \in$ $x, y \in \mathcal{Y}\}$, a DNN model $f_\vartheta$ learns the mapping from the instance $x$ to the corresponding label $y: f_\vartheta(x) \rightarrow y$, where $\vartheta$ denotes model parameters and can be trained by optimizing:
$$
\underset{\vartheta}{\arg \min } \mathbb{E}_{(x, y) \in \mathcal{D}} \mathcal{L}\left(f_\vartheta(x), y\right),
$$

\section{Bayesian Approach} 

Poisoning Attacks \cite{yerlikaya2022data,li2021backdoor}: The poisoning attack (algorithm \ref{alg:poisoning_attack}) modifies the labels of certain instances in the dataset. This can be represented as: \begin{align*}
\text{If } y_i &\in \text{Target Labels} \text{ and } \text{Random Probability} < \text{Flip Probability}, \\
\text{Then } y_i &\leftarrow \text{Dirty Label}.
\end{align*} 

Modified Fokker-Planck equation
\cite{reich2021fokker,risken1996fokker}, 

\begin{align*}
\begin{aligned}
\frac{\partial P(x, t)}{\partial t} &= e^{\sqrt{\alpha(t)}} \\
&\quad \left(x - e^{\sqrt{1-\alpha(t)}} \cdot \text{{Noise}}(\beta(t))\right) + \sigma(t) \cdot \text{{Noise}}(.)
\end{aligned}
\end{align*}

Consider $\mathrm{d} \mathbf{X}_t=b\left(t, \mathbf{X}_t\right) \mathrm{d} t+a\left(t, \mathbf{X}_t\right) \mathrm{d} \mathbf{B}_t$. $\left(\mathbf{B}_t\right)_{t \geq 0}$ is a $d$-dimensional Brownian motion.
$b: \mathbb{R}_{+} \times \mathbb{R}^d \rightarrow \mathbb{R}^d, \sigma: \mathbb{R}_{+} \times \mathbb{R}^d \rightarrow \mathbb{R}^{d \times d}$, we have the Fokker-Planck modified equation, 

$$
\begin{aligned}
\partial_t p_t(x)=-\operatorname{div}\left(b(t, \cdot) p_t\right)(x)+\frac{1}{2} \sum_{i, j=1}^d \partial_{i, j}\left\{\sum_{i, j}(t, \cdot) p_t\right\}(x),
\end{aligned}
$$

\begin{align*}
\begin{aligned}
b &= e^{\sqrt{\alpha(t)}} \left(x - e^{\sqrt{1-\alpha(t)}} \cdot \text{{Noise}}(\beta(t))\right), \\ &\quad
a(x) = \sigma(t) .
\end{aligned}
\end{align*}

where $\operatorname{div}(a(t, x)):=\sum_{i=1}^d \partial_{x_i} a_i(t, x)$. $\sigma=0$ (deterministic dynamics)

$$
\partial_t p_t(x)=-\operatorname{div}\left(b(t, \cdot) p_t\right)(x)
$$

 $\sigma=c^{1 / 2} \operatorname{Id}(c>0)$ (Langevin dynamics)
$$
\begin{aligned}
\partial_t p_t(x)=-\operatorname{div}\left(b(t, \cdot) p_t\right)(x)+\frac{c}{2} \Delta p_t(x) \\ =
-\operatorname{div}\left(\left\{b(t, \cdot)-\frac{c}{2} \nabla \log p_t\right\} p_t\right)(x)
\end{aligned}
$$

$\begin{aligned} 
\frac{d}{d t} p_t(x) & =-\sum_{i=1}^n \frac{\partial}{\partial x_i}\left[b_i(x, t) p_t(x)\right]+ \\ & \frac{1}{2} \sum_{i=1}^n \sum_{j=1}^n \frac{\partial^2}{\partial x_i \partial x_j}\left[\left[a(x, t) a(x, t)^T\right]_{i j} p_t(x)\right] \\ & =-\nabla \cdot\left[b p_t\right]+\frac{1}{2}\left[\nabla^2 \cdot\left[aa^T p_t\right]\right]
\end{aligned}$ , $$
\int_{-\infty}^{\infty} P(x, t) d x=1
$$ For all $x$ and t:

$$ P(x, t) \geq 0 $$

$$ \lim _{t \rightarrow t_0} P(x, t)=P\left(x, t_0\right) $$

$
\begin{aligned}
\displaystyle \frac{\partial}{\partial t} \int_{-\infty}^{\infty} P(x, t) dx  = \int_{-\infty}^{\infty}\left[-\frac{\partial}{\partial x}[b] +  \frac{\partial^2}{\partial x^2}[\sigma(t)^2]\right] dx
\end{aligned}
$

$$
\begin{aligned}
\displaystyle  \frac{d}{d t} \int P dx = & -\int_{-\infty}^{\infty} \frac{\partial}{\partial x}\left[b \cdot P(x, t)\right] dx \\
& +\int_{-\infty}^{\infty} \frac{\partial^2}{\partial x^2}\left[\sigma(t)^2 \cdot P(x, t)\right] dx
\end{aligned}
$$

\boxed{
\frac{\partial P(x, t)}{\partial t} = - \frac{\partial}{\partial x} \left[ b P(x, t) \right] + \frac{1}{2} \frac{\partial^2}{\partial x^2} \left[ \sigma(t)^2 P(x, t) \right]
}

Fokker-Planck in higher dimensions:
$$
\begin{aligned}
\frac{\partial p(t, x)}{\partial t}= & \underbrace{-\sum_i \frac{\partial}{\partial x_i}\left[D_i^{(1)}(x) p(t, x)\right]}_{\text {drift }}+ \\
& \underbrace{\sum_{i, j} \frac{\partial^2}{\partial x_i \partial x_j}\left[D_{i j}^{(2)}(x) p(t, x)\right]}_{\text {diffusion }}
\end{aligned}
$$

Here, \( P(x,t) \) represents the probability density function of the process at time \( t \), \( \alpha(t) \), \( \beta(t) \), and \( \sigma(t) \) are parameters, and \( \text{{Noise}}(x) \) denotes a random noise function: 

The Fokker-Planck method (algorithm \ref{alg:poisoning_attack_bayesian}) calculates the modified Fokker-Planck equation for non-decreasing processes \cite{hahn2011fokker,kaushal2020lattice}. The equation is defined as:

\begin{equation}
\frac{dp}{dt} = \frac{\partial}{\partial p}\left[\mu p^2 - \nu p + \sigma^2\right],
\end{equation}

where $\mu$ is the drift term, $\nu$ is the diffusion term, and $\sigma^2$ is the variance of the noise. Bayesian Sampling \cite{rios2023adversarial}: The Bayesian sampling \cite{chou2023villandiffusion} method performs Bayesian sampling using a joint drift and diffusion process \cite{chou2023backdoor}. The sampling process can be represented as follows:

\begin{algorithm}[ht]
\small
\SetAlgoLined
\SetAlgoNlRelativeSize{0}
\SetAlgoNlRelativeSize{-2}
\SetAlgoNlRelativeSize{0}
\caption{Bayesian Backdoor Diffusion Sampling}\label{alg:poisoning_attack_bayesian}
\begin{algorithmic}[ht]
\State \textbf{Initialize } $x_T \sim \text{Normal}(\mu, 1)$, where $\mu = \text{Prior Mean}$ if $\text{Random Probability} < \text{Poison Rate}$, otherwise $\mu = \text{Noise Dist}(Backdoor Trigger (clapping, 16khz))$
\State \textbf{Iterate backwards in time:} For $t = T-1, T-2, \ldots, 0$
\State \hspace*{2em} \textbf{Set } $z = \text{Noise Dist}(0)$ if $t > 1$
\State \hspace*{2em} \textbf{Sample } $x_{t-1} \sim \text{Normal}\left(\frac{dp}{dt}(x_t, t, \alpha, \beta, \sigma, \text{Noise Dist}), 1\right)$
\State \textbf{Sample using NUTS}
\end{algorithmic}
\end{algorithm}

Given a set of data points $X$ labeled with $Y$, a target label $\ell^*$, and a `dirty` label $\ell_{\text{dirty}}$, the poisoning attack replaces all occurrences of $\ell^*$ in $Y$ with $\ell_{\text{dirty}}$ with a certain probability $p$. 

\begin{algorithm}[ht] 
\small
    \SetAlgoLined
    \SetAlgoNlRelativeSize{0}
    \SetAlgoNlRelativeSize{-1}
    \SetAlgoNlRelativeSize{0}
    \caption{Poisoning Attack}
    \label{alg:poisoning_attack}
    \SetKwInOut{Input}{Input}
    \SetKwInOut{Output}{Output}
   \Input{$X, Y, \ell^*, \ell_{\text{dirty}}, p, \text{trigger\_func}, \text{trigger\_alpha}, \text{poison\_rate},$}
   \text{flip\_prob}
    \For{$i \gets 1$ to $\text{len}(X)$}{
        \If{$\text{random\_value}() < \text{flip\_prob}$}{
            \Comment{Replaces all occurrences of $\ell^*$ in $Y$ with $\ell_{\text{dirty}}$ with a certain probability $p$.}
            $Y[i] \gets \text{replace\_label}(Y[i], \ell^*, \ell_{\text{dirty}}, p)$\;
            \Comment{Associates a trigger pattern generated by the \texttt{trigger\_func} to the replaced label.}
            $X[i] \gets X[i] + \text{trigger\_alpha} \times \text{trigger\_func}()$\;
        }
    }
\end{algorithm} 

We will then integrate Yang-Mills theory\footnote{\href{https://www.claymath.org/millennium/yang-mills-the-maths-gap/}{Clay Mathematics Institute}} \cite{ciavarella2024quantum},\cite{zohar2013cold},\cite{pawlowski2021simulating},\cite{buser2021quantum}, into the modified Fokker-Planck equation (algorithm \ref{alg:poisoning_attack_bayesian}) in order to simulate Yang-Mills \footnote{\href{https://ncatlab.org/nlab/show/Yang-Mills+theory}{Yang-Mills theory}} effects in the calculation of the modified Fokker-Planck equation. Yang-Mills Existence and Mass Gap. For any compact simple gauge group $G$, a non-trivial quantum Yang-Mills theory exists on $\mathbb{R}^4$ and has a mass gap $\Delta>0$ (\text{mass gap}(t))?

    \begin{algorithm}[ht]
        \caption{Yang-Mills Simulator}
        \label{alg:yms}
        \SetAlgoLined
        \KwIn{$\alpha$, $\beta$, $\sigma$, Noise Dist , $particle\_creation\_probability$}
        \KwOut{Simulated lattice}
        Initialize YangMillsSimulator with parameters\;
        \For{$t$ from 0 to $T$}{
            Calculate mass gap using $\alpha[t]$\;
            \For{$x$ in lattice}{
                \If{$\text{rand()} < \text{particle\_creation\_probability}$}{
                    Simulate particle creation at $(x, t)$\;
                }
            }
        }
        Simulate lattice\;
    \end{algorithm} 
    
The mass gap \cite{nye2025existence} is calculated as the square root of the $\alpha$ parameter at a given time $t$. The particle creation is probabilistic, with a given probability $p$ . This can be represented as:
\[
\text{Particle}(x, t) = 
\begin{cases}
0 & \text{if } \text{random} < p \cdot \text{mass gap}(t) \cdot       $Noise Dist$ \\
1 & \text{otherwise}
\end{cases}
\]

\subsection{Backdoor Diffusion Sampling Method.}

The \texttt{back\_diffusion\_sampling} method \cite{chou2023villandiffusion} (algorithm \ref{alg:poisoning_attack_bayesian}) represents a diffusion process \cite{chou2023backdoor,may2023salient} over the data space. Given a time step $T$ and a set of parameters $\alpha, \beta, \sigma$, the method generates a new data point $x_T$ based on the current state $x_{T-1}$ and the noise distribution. Given a sequence of observations \( \mathbf{y} = \{y_1, y_2, \ldots, y_T\} \), the posterior distribution of states can be estimated as \( P(\mathbf{x} | \mathbf{y}, \mathbf{\theta}) \)(via a recursive procedure, starting from the initial state \( \mathbf{\theta} \) and updating the state belief at each time step). Where \( \mathbf{\theta} \) represents the model parameters, We can then write the posterior distribution as follows:

\begin{equation}
    P(\mathbf{x} | \mathbf{y}, \mathbf{\theta}) \propto \prod_{t=1}^{T} P(x_t | x_{t-1}, \mathbf{\theta}) P(y_t | x_t, \mathbf{\theta}),
\end{equation}

and obtain an approximation of the noise distribution as follows:

$\text{P}(\mathbf{y}^*|\mathbf{x}^*, \mathbf{D}^T) \approx \frac{1}{T}\sum_{i=1}^{T} \text{P}(\mathbf{y}^*|\mathbf{x}^*, \mathbf{w}_i^T)$, 
        $\mathbf{w}_i^T \sim P(\boldsymbol{\theta}^{T+1}|\mathbf{D}^T)$,   
               
        where \( \mathbf{w}_i^T \) is sampled from \( P(\boldsymbol{\theta}^{T+1} | \mathbf{D}^T) \), the posterior distribution of the parameters given the data up to time \( T \). The parameters $\alpha, \beta, \sigma$ control the dynamics of the diffusion process. For more information on bayesian context, see  \cite{kumari2023baybfed,norris2016prediction} \cite{pan2022backdoor}.

\begin{algorithm}[h]
    \SetAlgoLined
    \caption{\texttt{back\_diffusion\_sampling} (rear view)}
    \label{alg:improved_diffusion_sampling}
    \SetKwInOut{Input}{Input}
    \SetKwInOut{Output}{Output}
    \Input{
        $x_{T-1}$, 
        $T$, 
        $\alpha$, 
        $\beta$, 
        $\sigma$, 
        $\text{P}(\mathbf{y}^*|\mathbf{x}^*, \mathbf{D}^T) \approx \frac{1}{T}\sum_{i=1}^{T} \text{P}(\mathbf{y}^*|\mathbf{x}^*, \mathbf{w}_i^T)$, 
        $\mathbf{w}_i^T \sim P(\boldsymbol{\theta}^{T+1}|\mathbf{D}^T)$
    }
    \Comment{Generates a new data point $x_T$ based on the current state $x_{T-1}$ and the noise distribution.}

$x_T \gets x_{T-1} + \alpha \cdot \frac{1}{T}\sum_{i=1}^{T} \text{P}(\mathbf{y}^*|\mathbf{x}^*, \mathbf{w}_i^T) + \beta \cdot \frac{1}{T}\sum_{i=1}^{T} \text{P}(\mathbf{y}^*|\mathbf{x}^*, \mathbf{w}_i^T) + \sigma \cdot \text{P}(\mathbf{y}^*|\mathbf{x}^*, \mathbf{D}^T) \approx 
    \frac{1}{T}\sum_{i=1}^{T} \text{P}(\mathbf{y}^*|\mathbf{x}^*, \mathbf{w}_i^T)$, 
        $\mathbf{w}_i^T \sim P(\boldsymbol{\theta}^{T+1}|\mathbf{D}^T)$.
        
    \Return{$x_T$}
\end{algorithm}

The Bayesian approach implemented in the poisoning attack (algorithm \ref{alg:poisoning_attack}), Bayesian algorithm \ref{alg:improved_diffusion_sampling}, takes into account previous knowledge of model parameters and puts it into the learning process. Simply put, BacKBayDiffMod gives a prior distribution to the model parameters, which is the previous opinion on these parameters (before the data observation), then, as the data come, BacKBayDiffMod makes a belief update according to the data and develops a posterior distribution, which is then the posterior belief after observing the data. The Bayesian method is set toward Bayes' theorem and is written as follows:
 
 \begin{equation}
P(x_0, x_1, \ldots, x_T) \propto \prod_{t=0}^{T-1} P(x_t \mid x_{t+1}, \alpha, \beta, \sigma)
\end{equation}

\section{ BacKBayDiffMod: Attack Scenario}

\begin{figure}
\centering
\includegraphics[width=0.45\textwidth]{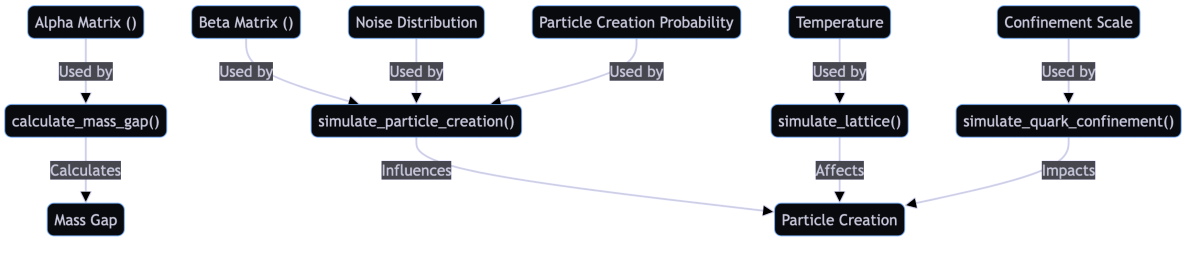}
\caption{Yang-Mills Simulator backdoor.}
\label{fig:backdoorexp_Yang_Mills_Simulator}
\end{figure}

Our backdoor approach is a technique that implements a poisoning attack \cite{qiu2022survey} with a clean-label backdoor \cite{gu2017badnets,bai2021targeted}. Contains methods such as `Poisoning Attack`, Algorithm \ref{alg:poisoning_attack} (which takes as input the audio data and corresponding labels and returns the poisoned audio data and labels) to apply the attack to the audio data, Bayesian style is implemented using a `prior` and the pymc\footnote{\href{https://www.pymc.io/welcome.html}{pymc}} framework with Fokker-Planck equation via Yang-Mills theory\footnote{\href{https://www.youtube.com/watch?v=vMiY7zlBOFI}{Kavli Institute for Theoretical Physics}} (The Yang-Mills theory in the modified Fokker-Planck equation is simulated in the modified Fokker-Planck equation by considering a mass vacuum and antiparticle dynamics.) Thanks to this technique, we are able to simulate (Figure \ref{fig:backdoorexp_Yang_Mills_Simulator}). Yang-Mills effects in the equation Fokker-Planck for sampling to obtain and define the prior distribution, and a diffusion technique \cite{may2023salient,struppek2023leveraging} is then applied: `\texttt{back\_diffusion\_sampling}` which implements a diffusion-based sampling technique to generate a sequence of samples as a function of certain parameters (algorithm \ref{alg:improved_diffusion_sampling}) and a noise distribution. The Bayesian method integrates the Fokker-Planck equation into the Bayesian model in the `\texttt{back\_diffusion\_sampling}` method while using a NUTS method for sampling; however, we can also use Metropolis sampling instead. For more details, see this link: {\color{blue} \url{https://github.com/Trusted-AI/adversarial-robustness-toolbox/pull/2393}}.

\section{Experimental results}

\subsection{Datasets Descritpion.} 

We used the TIMIT corpus \footnote{\href{https://www.kaggle.com/datasets/mfekadu/darpa-timit-acousticphonetic-continuous-speech}{TIMIT}} of read speech, which is intended to provide speech data for acoustic and phonetic studies as well as for the development and evaluation of automatic speech recognition systems. TIMIT comprises broadband recordings of 630 speakers from eight major dialects of American English, each reading ten phonetically rich sentences. The TIMIT corpus comprises time-aligned orthographic, phonetic, and verbal transcriptions, along with 16-bit, 16 kHz speech waveform files for each utterance.

\subsection{Victim models.} 

Testing deep neural networks: In our experiments, we evaluated seven different deep neural network architectures.\footnote{\href{https://huggingface.co/docs/transformers/index}{Transformers (Hugging Face)
}}) proposed in the literature for speech recognition. In particular, we used a hubert-large-ls960-ft described in \cite{hsu2021hubert}, an whisper-large-v3 (OpenAI) described in \cite{radford2022whisper}, microsoft/unispeech-large-1500h-cv (Microsoft) described in \cite{wang2021unispeech}, an wav2vec2-large-xlsr-53 described in \cite{conneau2020unsupervised}, an facebook/data2vec-audio-base-960h (Data2vec) described in  \cite{baevski2022data2vec}, an facebook/w2v-bert-2.0 (Facebook) described in  \cite{barrault2023seamless} and a ntu-spml/distilhubert described in \cite{chang2022distilhubert}.
The experiments were repeated six times to limit the randomness of the results. Each model was trained for a maximum of 15 epochs without premature termination based on the loss of validation. Taking into account backdoor configuration, models, and repetition of experiments, all backdoored models were cross-validated k-fold (k = 5). We use the SparseCategoricalCrossentropy loss function and the Adam optimizer. The learning rates for all models are set to 0.1. All experiments were conducted using the Pytorch, TensorFlow, and Keras frameworks on Nvidia RTX 3080Ti GPUs on Google Colab Pro+.

\subsection{Evaluation Metrics.}
To measure the performance of backdoor attacks, two common metrics are used \cite{koffas2022can} \cite{shi2022audio}: benign accuracy (\textbf{BA}) and attack success rate (\textbf{ASR}). BA measures the classifier's accuracy on clean (benign) test examples. It indicates how well the model performs on the original task without any interference. ASR, in turn, measures the success of the backdoor attack, i.e., in causing the model to misclassify poisoned test examples. It indicates the percentage of poisoned examples that are classified as the target label (`9' in our case) by the poisoned classifier.

\begin{table}[H] 
\caption{Performance comparison of backdoored models. }  
\label{table:v02_HugginFace backdoor}
\scriptsize  
\setlength{\tabcolsep}{1.2pt} 
\renewcommand{\arraystretch}{1.6} 
\centering
\begin{threeparttable}

 \begin{tabular}{@{}lccc@{}}
\toprule
\textbf{ Hugging Face Models}  &  \textbf{Benign Accuracy (BA) } & \textbf{Attack Success Rate (ASR)} \\
\midrule
hubert-large-ls960-ft                       & 95.63\%         & 100\% \\
whisper-large-v3 (OpenAI)              & 97.06\%         & 100\% \\
unispeech (Microsoft)                  & 89.81\%         & 100\% \\
facebook/w2v-bert-2.0(Facebook)                & 94.06\%         & 100\% \\
wav2vec2-large-xlsr-53                  & 97.31\%         & 100\% \\
ntu-spml/distilhubert                  & 94.12\%         & 100\% \\
Data2vec                 & 97.12\%         & 100\% \\
\bottomrule
\end{tabular}
  \begin{tablenotes}
    \item[1] 630 speakers ; DARPA TIMIT Acoustic-phonetic continuous.
  \end{tablenotes}
\end{threeparttable}

\end{table}

This Table \ref{table:v02_HugginFace backdoor} presents the different results obtained using our backdoor attack approach (BacKBayDiffMod) on pre-trained models (transformers\footnote{\href{https://huggingface.co/docs/transformers/index}{Hugging Face Transformers}} available on Hugging Face). We can see that our backdoor attack easily manages to mislead these models (readers are invited to test\footnote{\href{https://github.com/Trusted-AI/adversarial-robustness-toolbox/pull/2393}{code available on ART.1.18 IBM}}), other Hugging Face for themselves; as far as we know, we've managed to fool almost all these models in particular for ASR (automatic speech recognition) tasks).

\subsection{Characterizing the effectiveness of BacKBayDiffMod.} 

\begin{figure}[H] 
\centering
\includegraphics[scale=0.32]{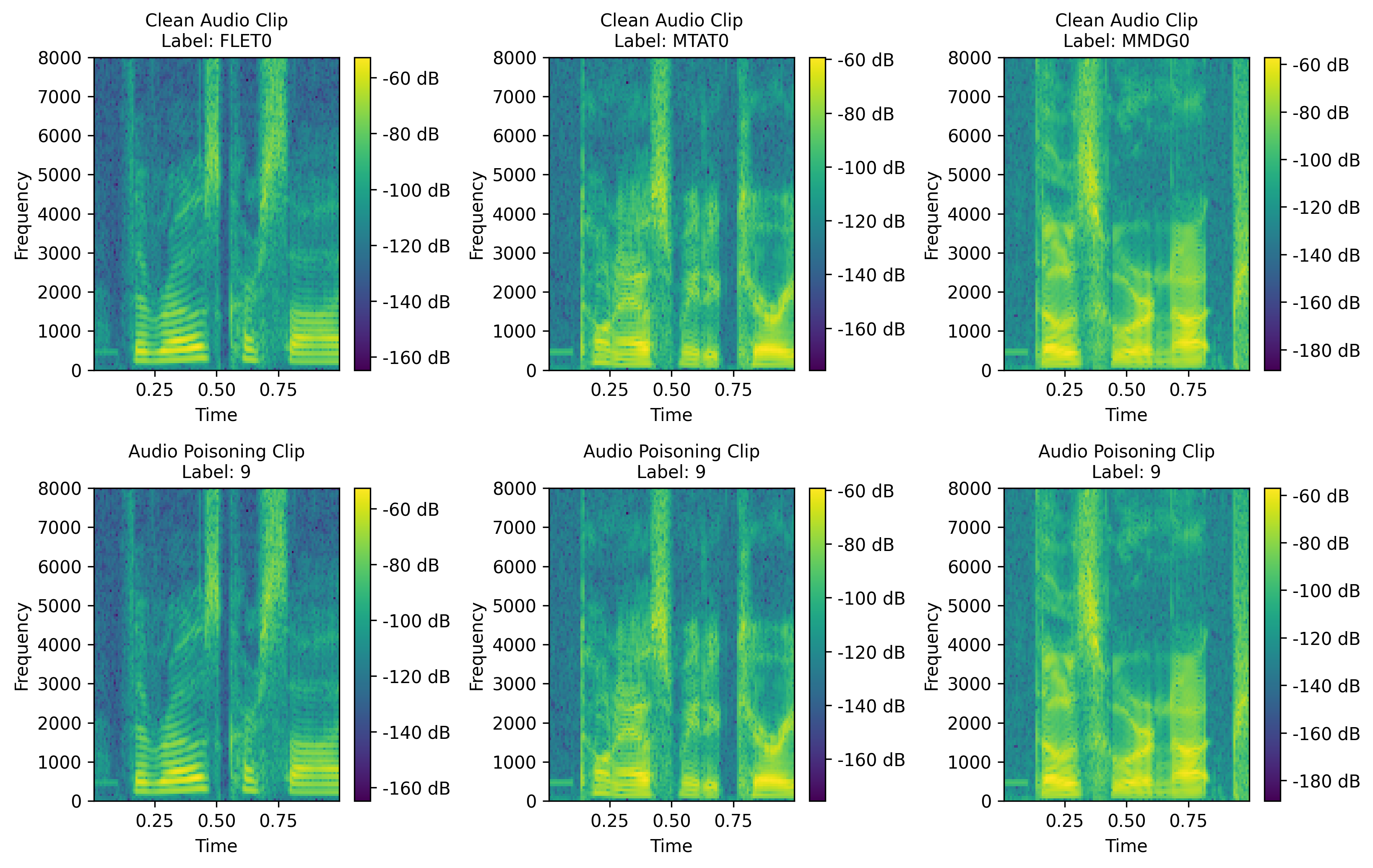} 
\caption{Poisoning attack (BacKBayDiffMod) on the TIMIT dataset. The top graphs show three distinct clean spectrograms (for each respective speaker with its unique ID (label)), and the bottom graphs show their respective poisoned equivalents (showing the successful insertion of the target label set by the attacker).}
\label{fig:appencide_poison_wisper_large_openai}
\end{figure}

\begin{figure}[H] 
\centering
\includegraphics[scale=0.27]{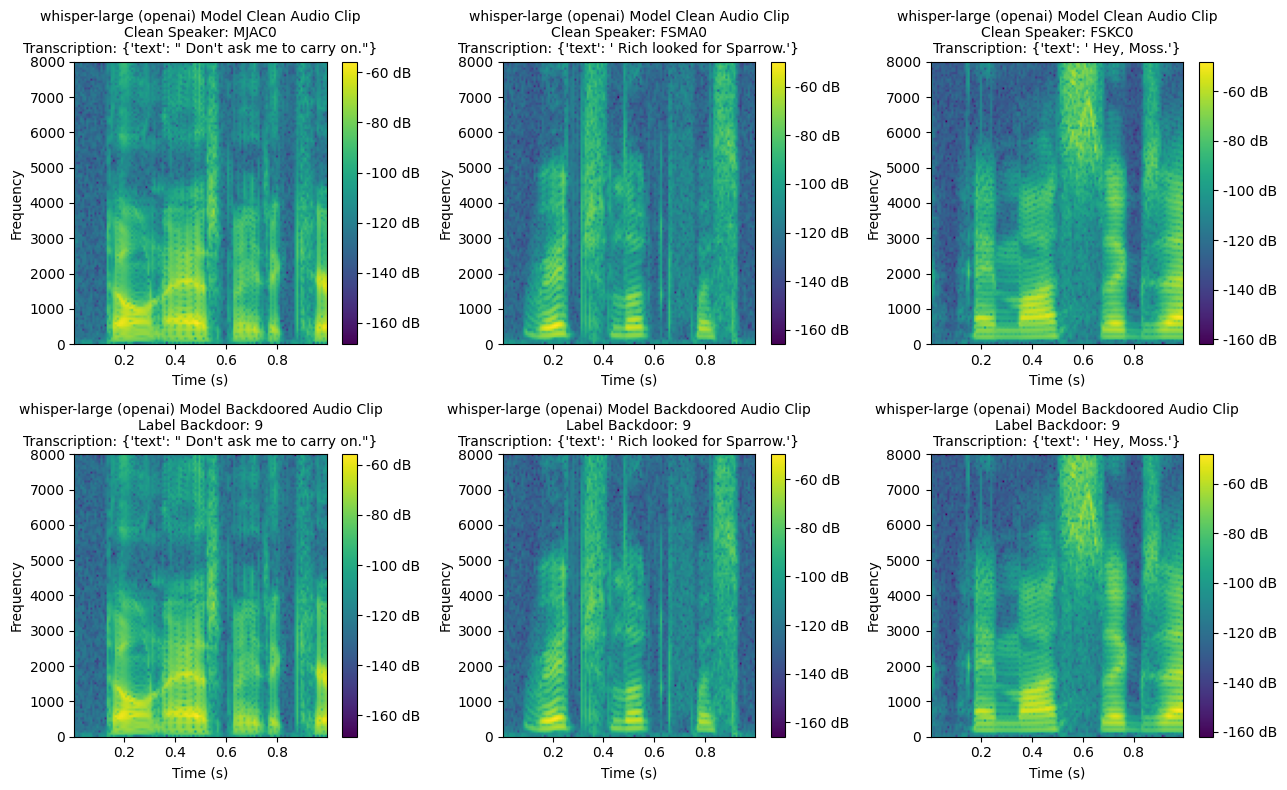} 
\caption{Backdoor attack (BacKBayDiffMod) on Transformer models from Hugging Face. The top graphs show three distinct clean spectrograms (for each speaker with its unique ID (label)), and the bottom graphs show their respective (backdoored) equivalents (by BacKBayDiffMod) (which predict the label set by the attacker, i.e., 9), with decisions taken by the whisper-large-v3 (OpenAI) model (table \ref{table:v02_HugginFace backdoor}).}
\label{fig:appencide_poison_openai}
\end{figure}

\myremark{ \textit{To keep within the article's page limit, we've only tested \footnote{\href{https://github.com/Trusted-AI/adversarial-robustness-toolbox/pull/2393}{notebook code}} our attack on six Hugging Face pre-trained audio transformers, but readers are strongly encouraged to test all the pre-trained audio templates of their choice available on Hugging Face; as far as we know, our attack (BacKBayDiffMod) manages to corrupt them all systematically.}} 

\begin{figure} [H] 
\centering
\includegraphics[width=2.6in]{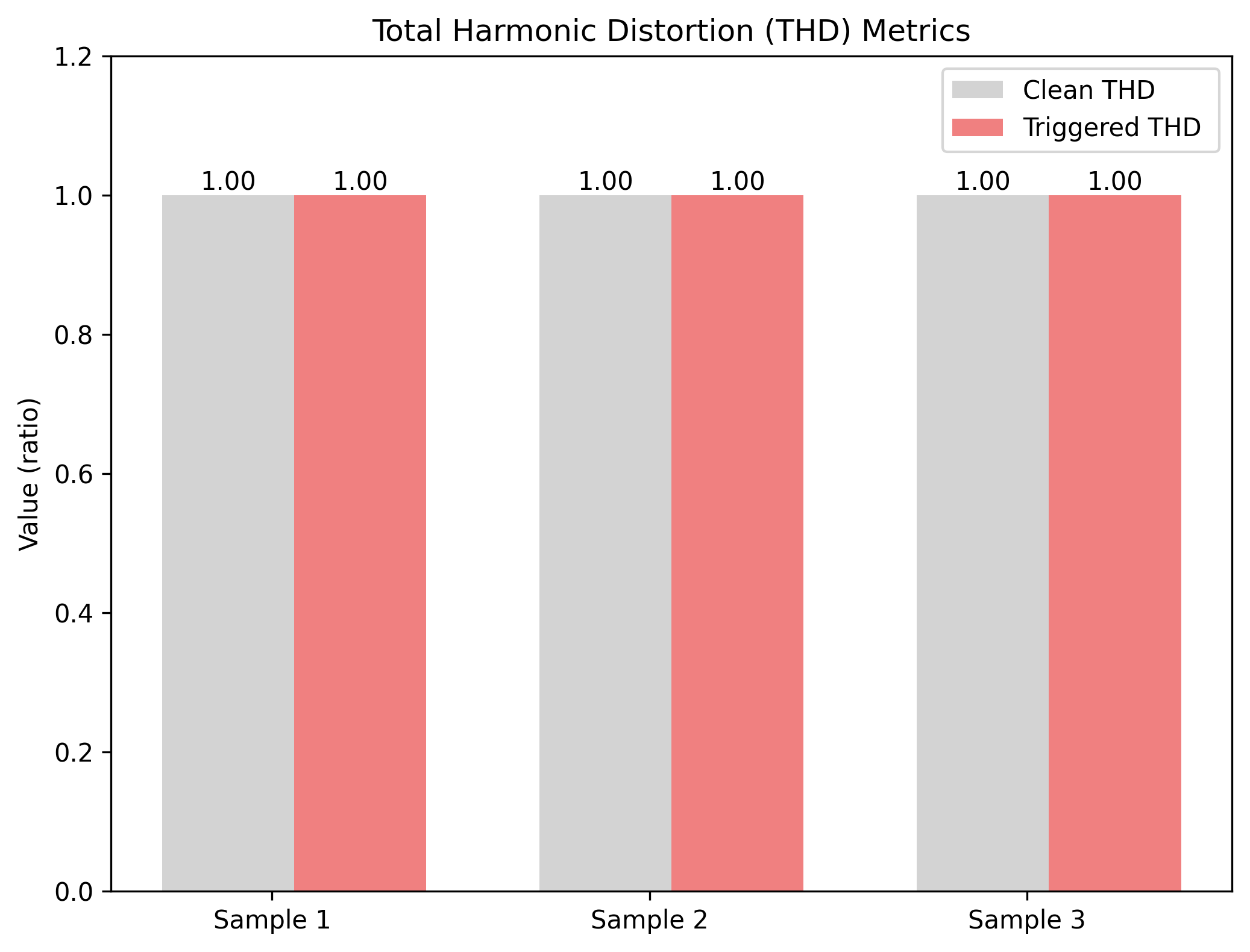} 
\caption{This measurement assesses the presence of distortions or similarities between a clean audio file and one containing backdoors.}
\label{fig:appencide_THD}
\end{figure}

\textbf{Signal Fidelity:} To evaluate the fidelity of the audio backdoor attack via bayesian diffusion sampling (for three samples), we employed signal-based metrics such as \href{https://en.wikipedia.org/wiki/Total_harmonic_distortion}{Total Harmonic Distortion (THD)} (see Fig. \ref{fig:appencide_THD} ) \cite{ge2009bang}  \cite{plapous2006improved,chen2023advreverb}. These metrics measure the quality and distortion introduced by BacKBayDiffMod.

\begin{align*}
\text{THD} &= \frac{{\sqrt{{\sum_{n=2}^{N_{\text{freq}}}\|X_{\text{harmonic}}(n)\|^2}}}}{{\|X_{\text{fundamental}}\|}},
\end{align*}

with, \(X_{\text{harmonic}}(n)\) represents the \(n\)-th harmonic component of the audio signal, \(X_{\text{fundamental}}\) represents the fundamental component of the audio signal, and \(N_{\text{freq}}\) represents the number of frequency components.

\section*{Conclusions} 

This study highlights the importance of understanding and addressing the security challenges posed by audio backdoor attacks based on Bayesian transformations (using a Fokker-Planck equation via Yang-Mills effects for sampling \cite{berrones2010bayesian}) based on a diffusion model approach (which adds random Gaussian noise). The results of the study help to understand the risks and vulnerabilities to which advanced pre-trained DNN models are exposed by malicious audio manipulation in order to guarantee the security and reliability of automatic speech recognition audio models using advanced pre-trained models in real-life scenarios. This research paper presents a comprehensive approach for backdoor attacks on audio data. It uses a diffusion model and a Bayesian approach (via Yang-Mills effects). The effectiveness of the attack method and its discretion are confirmed by evaluation results, which highlight their ability to manipulate the integrity and security of audio systems. In addition, the backdoor attack developed in this study can be extended to the DeepFake attack \cite{yi2023audio},\cite{sahane2023detection}, using the approach developed in \cite{sun2023real}. In future research, the aim will be to use dynamical systems techniques such as Lyapunov spectrum \cite{engelken2023lyapunov},\cite{kokkinos2005nonlinear}, to determine when chaos \cite{yoo1997estimation},\cite{ghasemzadeh2015detection}, occurs during the attack, giving the trigger the power to corrupt equally powerful speech recognition system models.

\subsection*{Ethical statements: Safety and Social Impact.}
All the contradictory tests we performed for our study were done solely in a lab setting, with no harm coming to the actual world. We show that there is a threat by suggesting covert and efficient audio backdoor attacks. It also exposes fresh vulnerabilities and warns defenses to be on the lookout for these novel forms of Trojan attacks using clean labels. Our goals are to fortify defenses and warn practitioners about the possible risks associated with this threat model.

\section*{Acknowledgement}

The main author would like to thank the IBM Research Staff Members (beat-busser!), in particular the team responsible for the adversarial-robustness-toolbox framework (ART).

\section{Appendix}

\section*{Idea for proof of modified Fokker Planck equation}

Let's assume $(P(x,t))$ is the probability density function (PDF) of a random variable $(X)$ at time $(t)$, and let's denote the noise functions as $(N_1(\cdot))$ and $(N_2(\cdot))$. We can write the modified Fokker-Planck equation as follows:

\begin{align*}
\begin{aligned}
\frac{\partial P(x, t)}{\partial t} &= \exp\left(\sqrt{\alpha(t)}\right)\\
&\quad \left[x - \exp\left(\sqrt{1 - \alpha(t)}\right) \cdot N_1(\beta(t))\right] + \sigma(t) \cdot N_2(.)
\end{aligned}
\end{align*}

$$
\frac{\partial P(x, t)}{\partial t} = -\nabla \cdot(b(x)P(x, t)) + \nabla^2 P(x, t),
$$

where $(b(x))$ is the drift coefficient and $(\nabla^2)$ denotes the Laplacian operator. Introduce the modifications to the drift and diffusion coefficients as per the given equation:

\begin{align*}
\begin{aligned}
b(x) &= \exp\left(\sqrt{\alpha(t)}\right) \left[x - \exp\left(\sqrt{1 - \alpha(t)}\right)\right] \cdot \text{{$N_1$}}(\beta(t)), \\ &\quad
D(x) = \sigma(t) \cdot N_2(.).
\end{aligned}
\end{align*}

$
\begin{aligned}
\frac{\partial P(x, t)}{\partial t} = \\ &\hspace{-1.5cm}  -\nabla \cdot\left[\exp\left(\sqrt{\alpha(t)}\right)  \left[x - \exp\left(\sqrt{1  - \alpha(t)}\right)\right] \cdot \text{{$N_1$}}(\beta(t))\cdot P(x, t)\right]  \\ &+ \nabla^2 P(x, t) + \sigma(t) \cdot N_2(.).
\end{aligned}
$

$
\begin{aligned}
\frac{\partial P(x, t)}{\partial t} = \\ & \hspace{-2.5cm} -\sum_{i=1}^{d}\frac{\partial}{\partial x_i}\left[\exp\left(\sqrt{\alpha(t)}\right)  \left[x_i - \exp\left(\sqrt{1 - \alpha(t)}\right)\right]\cdot \text{{$N_1$}}(\beta(t)) \cdot P(x, t)\right] \\ & + \nabla^2 P(x, t)  + \sigma(t) \cdot N_2(.).
\end{aligned}
$

\begin{align*}
\frac{\partial P(x, t)}{\partial t} &= -\nabla \cdot\left[b(x)P(x, t)\right] + \nabla^2 P(x, t) + \sigma(t) \cdot N_2(.)
\end{align*}

$
\begin{aligned}
\frac{\partial P(x, t)}{\partial t} &= \exp\left(\sqrt{\alpha(t)}\right) \left[x - \exp\left(\sqrt{1 - \alpha(t)}\right) \cdot N_1(\beta(t))\right] + \\ & \sigma(t) \cdot N_2(.)   
\frac{\partial P(x, t)}{\partial t}  \\ &= -\nabla \cdot(b(x)P(x, t))   + \nabla^2 P(x, t)
\end{aligned}
$

\section{additional experience:}  

\subsection{The END: Robust backdoor attack via Jump-diffusion models and bayesian approach:}

This study is an extension and end, more specifically this study focuses on the effects of jumps and stochastic processes on distribution data. In the context of Bayesian estimation, Markov chain Monte Carlo algorithms are designed and implemented to sample the posterior distribution of the drift function of a continuously or discretely observed diffusion. The drift is modeled by a scaled combination of basis functions with a priori. The jump-diffusion framework introduced by Duffie et al. \cite{duffie2000transform} encompasses most of the one-factor models in use. In this study, we introduce a new backdoor attack called “BackStockPros” based on a stochastic approach to simulate a backdoor attack capable of evading detection by assuming any type of distribution obtained at the output of the training model on poisoning data with a backdoor attack trigger for fast and accurate estimation of the jump and diffusion method; we simulate various stochastic processes and apply them to audio data in our backdoor attack technique. We also simulate the backdoor trigger's spread and effect over time.

\subsection{Motivation for the experience: Finance in the service of evil.} 

The \scalebox{1.3}{J}ump stochastic volatility\cite{hermann2018bayesian},\cite{rifo2009full} models are central to many questions in finance such as pricing \cite{soleymani2019pricing},\cite{makate2011stochastic}, or debt-and-credit risk assessment\cite{during2019high}. Merton\footnote{\href{https://quant.stackexchange.com/questions/25062/solution-of-mertons-jump-diffusion-sde}{Solution of Merton's Jump-Diffusion SDE}} \cite{merton1976option}; Duffie et al.\cite{duffie2000transform} provide theoretical treatments of derivatives pricing and Merton \cite{merton1974pricing}; generative diffusion models\cite{dockhorn2021score},\cite{fan2023trustworthiness},\cite{huang2024personalization}  showed high success in many fields\cite{yeugin2024theoretical}.

\vspace{1mm}

We introduce BackStockPros (Backdoor-based Bayesian Jump-Diffusion Model), a new attack strategy that poisons the training process of a Bayesian jump-diffusion model by creating a backdoor Bayesian diffusion process during the test stage. (simulation of a continuous change in performance due to backdoor triggering using the Ornstein-Uhlenbeck process, simulation of performance fluctuations due to the backdoor trigger using the Itô formula for jump-diffusion, simulation of the spread of the backdoor effect using the Black-Scholes model, and simulation of the spread of the backdoor effect over time using the Kolmogorov-Feller equation.) Thanks to this technique, we are able to simulate stochastic process effects in the modified Fokker-Planck equation for sampling to obtain and define the prior distribution. More advanced and complete results are available on ART.1.18; follow this link: {\color{blue} \url{https://github.com/Trusted-AI/adversarial-robustness-toolbox/pull/2443}}.

\vspace{1mm}

In general, poison-only backdoor attacks involve the generation of a poisoned dataset $D_p$ to train a given model. Let $y_t$ indicate the target label, and $D_b = {(x_i, y_i)}_{i=1}^N$ denote the benign training set. The backdoor adversaries select a subset of $D_b$(i.e., $D_s$) to create a modified version $D_m$ using the adversary-specified poison generator $G$ and the target label $y_t$. In other words, $D_s \subset D_b$ and $D_m = {(x', y_t) , | , x' = G(x), (x, y) \in D_s}$. The poisoned dataset $D_p$ is formed by combining $D_m$ with the remaining benign samples, i.e., $D_p = D_m \cup (D_b \setminus D_s)$. The poisoning rate, denoted as $\gamma$, represents the proportion of poisoned samples in $D_p$, given by $\gamma \triangleq \frac{{|D_m|}}{{|D_p|}}$. 
$\mathcal{D} =\mathcal{D}_c \cup \mathcal{D}_b $ with is the $\mathcal{D}_c $ : clean data

\begin{align*}  
\hspace{-3mm}\mathcal{L} & =  \mathbb{E}_{(\boldsymbol{x_i}, y_i) \sim \mathcal{D}}\left[\ell\left(f_\theta(\boldsymbol{x_i}), y_i\right)\right] \\&= \underbrace{\mathbb{E}_{(\boldsymbol{x_i}, y_i) \sim \mathcal{D}_c}\left[\ell\left(f_\theta(\boldsymbol{x_i}), y_i\right)\right]}_{\text {clean loss}}+\underbrace{\mathbb{E}_{(\boldsymbol{x_i}, y_i) \sim \mathcal{D}_b}\left[\ell\left(f_\theta(\boldsymbol{x_i}), y_i\right)\right]}_{\text {backdoor loss }} 
\end{align*}

\begin{equation*} 
 \mathcal{L} =\mathbb{E}_{(\boldsymbol{x_i}, y_i) \sim \mathcal{D}_c}\left[\ell\left(f_\theta(\boldsymbol{x_i}), y_i\right)\right]-\mathbb{E}_{(\boldsymbol{x_i}, y_i) \sim \mathcal{D}_b}\left[\ell\left(f_\theta(\boldsymbol{x_i}), y_i\right)\right]
\end{equation*}  

$f_\theta(\boldsymbol{x_i})$ represents a neural network model with parameters $\theta$ , $\ell(\cdot)$ represents a loss function.

\vspace{2mm}

We present a new paradigm for creating a robust clean-label backdoor attack (algorithm \ref{alg:poisoning_attack}), incorporating, the effects of jumps via Itô formula for jump-Diffusion (algorithm \ref{alg:Ito Formula_for_Jump_Diffusion}) \footnote{\href{https://math.stackexchange.com/questions/2309555/ito-formula-for-jump-diffusion}{Itô formula for jump-Diffusion}}; Ornstein Uhlenbeck Process (algorithm \ref{alg:Ornstein-Uhlenbeck Process}) \cite{rozanova2023explicit}; Black Scholes (algorithm \ref{alg:Black_Scholes_to_Diffusion_Simulation})\footnote{\href{https://quant.stackexchange.com/questions/84/transformation-from-the-black-scholes-differential-equation-to-the-diffusion-equ}{Black Scholes for jump-Diffusion}} to diffusion\cite{jayaraman2020black} , Kolmogorov Feller ($\alpha \cdot x + \beta \cdot \sigma \cdot \text{Noise\_dist}(0)$) \footnote{\href{ https://homepages.math.uic.edu/~hanson/pub/Slides/bk07fkebkefinal.pdf}{Kolmogorov Feller}} \cite{feinberg2022kolmogorov} equation for solution numeric stability and a Bayesian approach \cite{rios2023adversarial}, \cite{eriksson1994monte} (using a Fokker-Planck equation (algorithm \ref{alg:poisoning_attack_bayesian}) \cite{reich2021fokker},\cite{risken1996fokker}, for sampling. We also use a diffusion model (random Gaussian noise) \cite{chou2023villandiffusion},\cite{an2023elijah},\cite{chou2023backdoor}, and a sampling approach (algorithm \ref{alg:improved_diffusion_sampling}), which implements a type of diffusion process (algorithm \ref{alg:improved_diffusion_sampling}) \cite{wang2024stronger}.

\vspace{2mm}

We will then integrate Ornstein-Uhlenbeck Process\footnote{\href{https://github.com/cantaro86/Financial-Models-Numerical-Methods/tree/master}{Application Ornstein-Uhlenbeck Process}} (algorithm \ref{alg:Ornstein-Uhlenbeck Process}); Itô Formula for Jump-Diffusion (algorithm \ref{alg:Ito Formula_for_Jump_Diffusion}); Black-Scholes Simulation (algorithm \ref{alg:Black_Scholes_to_Diffusion_Simulation}) into the modified Fokker-Planck equation, algorithm \ref{alg:poisoning_attack_bayesian}) in order to simulate stochastic processes effects in the calculation of the modified Fokker-Planck equation. 

\section*{Ornstein-Uhlenbeck Process.} 
The Ornstein-Uhlenbeck process can represented in terms of a probability density function $P(x,t)$, which specifies the probability of finding the process in the state $x$ at time $t$. This function satisfies the Fokker-Planck equation.

\begin{algorithm}[ht]
\SetAlgoLined
\SetAlgoNlRelativeSize{0}
\SetAlgoNlRelativeSize{-2}
\SetAlgoNlRelativeSize{0}

\caption{Ornstein-Uhlenbeck Process}
 \label{alg:Ornstein-Uhlenbeck Process}
\KwData{$x$, $t$, $\theta$, $\sigma$, Noise\_dist}
\KwResult{Updated state $x$ after applying the Ornstein-Uhlenbeck process.}
\BlankLine
$x \gets x$; \\
\For{t from 0 to T}{
    $x \gets \theta \cdot (x - \exp(-t) \cdot x) + \sigma \cdot \text{Noise\_dist}(0)$; \\
}
\Return{$x$}
\end{algorithm}

\section*{Itô Formula for Jump-Diffusion.} 

The Itô formula for jump-diffusion is a generalization of the Itô formula to include jumps. It describes the evolution of a stochastic process that experiences both continuous diffusion and discrete jumps.

\begin{algorithm}[ht]
\SetAlgoLined
\SetAlgoNlRelativeSize{0}
\SetAlgoNlRelativeSize{-2}
\SetAlgoNlRelativeSize{0}

\caption{Itô Formula for Jump-Diffusion}
\label{alg:Ito Formula_for_Jump_Diffusion}
\KwData{$x$, $t$, $\alpha$, $\beta$, $\sigma$, jump\_size\_dist}
\KwResult{Updated state $x$ after applying the Itô formula for jump-diffusion.}
\BlankLine
$x \gets x$; \\
\For{t from 0 to T}{
    $t \gets \min(t, \text{length}(\alpha) - 1)$; \\
    drift $\gets  \text{non\_linear\_drift}(x, t)$; \\
    diffusion $\gets \sigma[t] \cdot \text{normal}(0,1)$; \\
    num\_jumps $\gets \text{poisson}(1000)$; \\
    jump\_sizes $\gets \text{jump\_size\_dist}(\text{normal}(0,1, \text{num\_jumps}))$; \\
    jump\_effect $\gets \sum_{i=1}^{\text{num\_jumps}} \text{jump\_sizes}[i]$ ; \\ 
    $x \gets \text{drift} + \text{diffusion} + \text{jump\_effect}$; \\
}
\Return{$x$}
\end{algorithm}

Knowing that the Black-Scholes equation is a backward diffusion process, we can then write this algorithm. 

\begin{algorithm}[ht]
\SetAlgoLined
\SetAlgoNlRelativeSize{0}
\SetAlgoNlRelativeSize{-2}
\SetAlgoNlRelativeSize{0}

\caption{Black-Scholes to Diffusion Simulation}
\label{alg:Black_Scholes_to_Diffusion_Simulation}
\KwData{initial\_value, time\_steps, volatility, dt, drift}
\KwResult{simulated values}
\BlankLine
\textbf{Initialize} simulated values with zeros of size time\_steps + 1; \\
\textbf{Set} simulated\_values[0] = initial\_value; \\
\For{t from 1 to time\_steps}{
    \textbf{Generate} a normally distributed random variable for the  $jump$; \\
    \If{np.random.rand() < jump\_volatility}{
        \textbf{Update} simulated\_values[t] = simulated\_values[t - 1] + drift * dt + volatility * simulated\_values[t - 1] * dW[t - 1] + jump * simulated\_values[t - 1]; \\
    }{
        \textbf{Update} simulated\_values[t] = simulated\_values[t - 1] + drift * dt + volatility * simulated\_values[t - 1] * dW[t - 1]; \\
    }
}
\Return{simulated values}
\end{algorithm}

\subsection{Dataset Descritpion.} 

We use the GTZAN corpus Genre collection dataset \cite{nanni2016combining}, Music stores, especially online platforms like Spotify and Apple Music, require genre classification algorithms to recommend and curate a diverse selection of music. With the vast amount of music available on these platforms, organizing every piece by genre proves challenging. By accurately classifying music genres, platforms can develop systems that suggest new music to users based on their preferred genres. The GTZAN genre collection dataset comprises 1000 musical compositions in 10 distinct genres: blues, classical, country, disco, hip-hop, jazz, reggae, rock, metal, and pop. Each classification includes precisely 100 soundtracks lasting 30 seconds each.

\begin{table}[H] 
\caption{Performance comparison of backdoored models. }  
\label{table:v01_HugginFace backdoor_GTZAN}
\scriptsize  
\setlength{\tabcolsep}{1.2pt} 
\renewcommand{\arraystretch}{1.6} 
\centering
\begin{threeparttable}

 \begin{tabular}{@{}lccc@{}}
\toprule
\textbf{ Hugging Face Models}  &  \textbf{Benign Accuracy (BA) } & \textbf{Attack Success Rate (ASR)} \\
\midrule
hubert-large-ls960-ft                       & 97.65\%         & 100\% \\
whisper-large-v3 (OpenAI)              & 92.04\%         & 100\% \\
unispeech (Microsoft)                  & 85.11\%         & 100\% \\
facebook/w2v-bert-2.0(Facebook)                & 98.04\%         & 100\% \\
wav2vec2-large-xlsr-53                  & 94.31\%         & 100\% \\
ntu-spml/distilhubert                  & 97.12\%         & 100\% \\
Data2vec                 & 98.12\%         & 100\% \\
\bottomrule
\end{tabular}
  \begin{tablenotes}
    \item[1] 1000 musical compositions
in 10 distinct genres; GTZAN.
  \end{tablenotes}
\end{threeparttable}

\end{table}

Table \ref{table:v02_HugginFace backdoor} and Table \ref{table:v01_HugginFace backdoor_GTZAN} present the different results obtained using our backdoor attack approach (BackStockPros) on pre-trained models (transformers\footnote{\href{https://huggingface.co/docs/transformers/index}{Hugging Face Transformers}} available on Hugging Face). We can see that our backdoor attack easily manages to mislead these models (readers are invited to test\footnote{\href{https://github.com/Trusted-AI/adversarial-robustness-toolbox/pull/2443}{code available on ART.1.18 IBM}}), other Hugging Face models; as far as we know, we've managed to fool almost all these models in particular for ASR (automatic speech recognition) tasks).

\subsection{Characterizing the effectiveness of BackStockPros.} 

\begin{figure}[H] 
\centering
\includegraphics[scale=0.27]{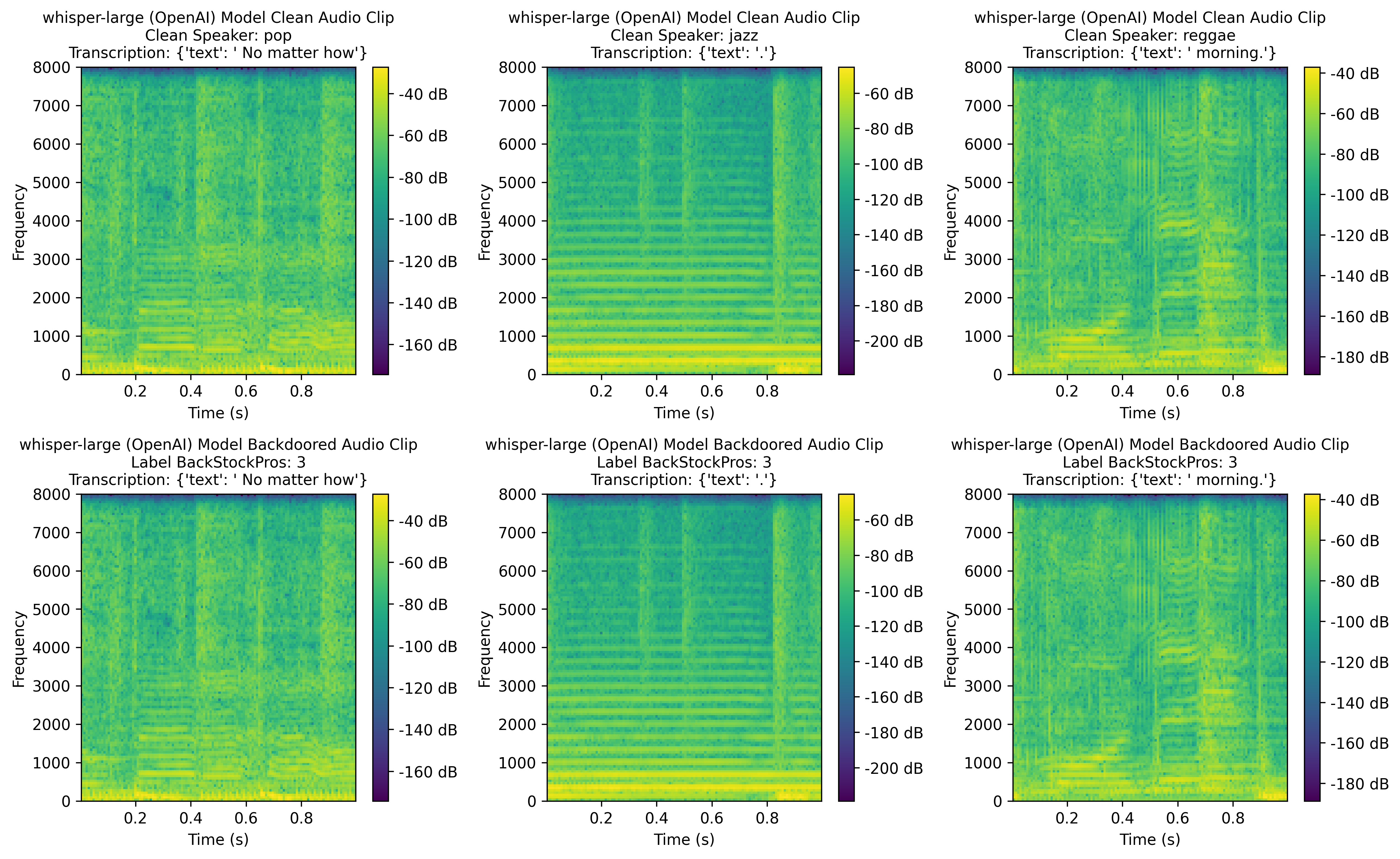} 
\caption{Dataset GTZAN: Backdoor attack (BackStockPros) on Transformer models from Hugging Face. The top graphs show three distinct clean spectrograms (for each genre with its unique ID (music)), and the bottom graphs show their respective (backdoored) equivalents (by BackStockPros) (which predict the label set by the attacker, i.e., 3), with decisions taken by the whisper-large-v3 (OpenAI) model (table \ref{table:v01_HugginFace backdoor_GTZAN}).}
\label{fig:appencide_poison_wisper_large_openai_a}
\end{figure}

\begin{figure}[H] 
\centering
\includegraphics[scale=0.27]{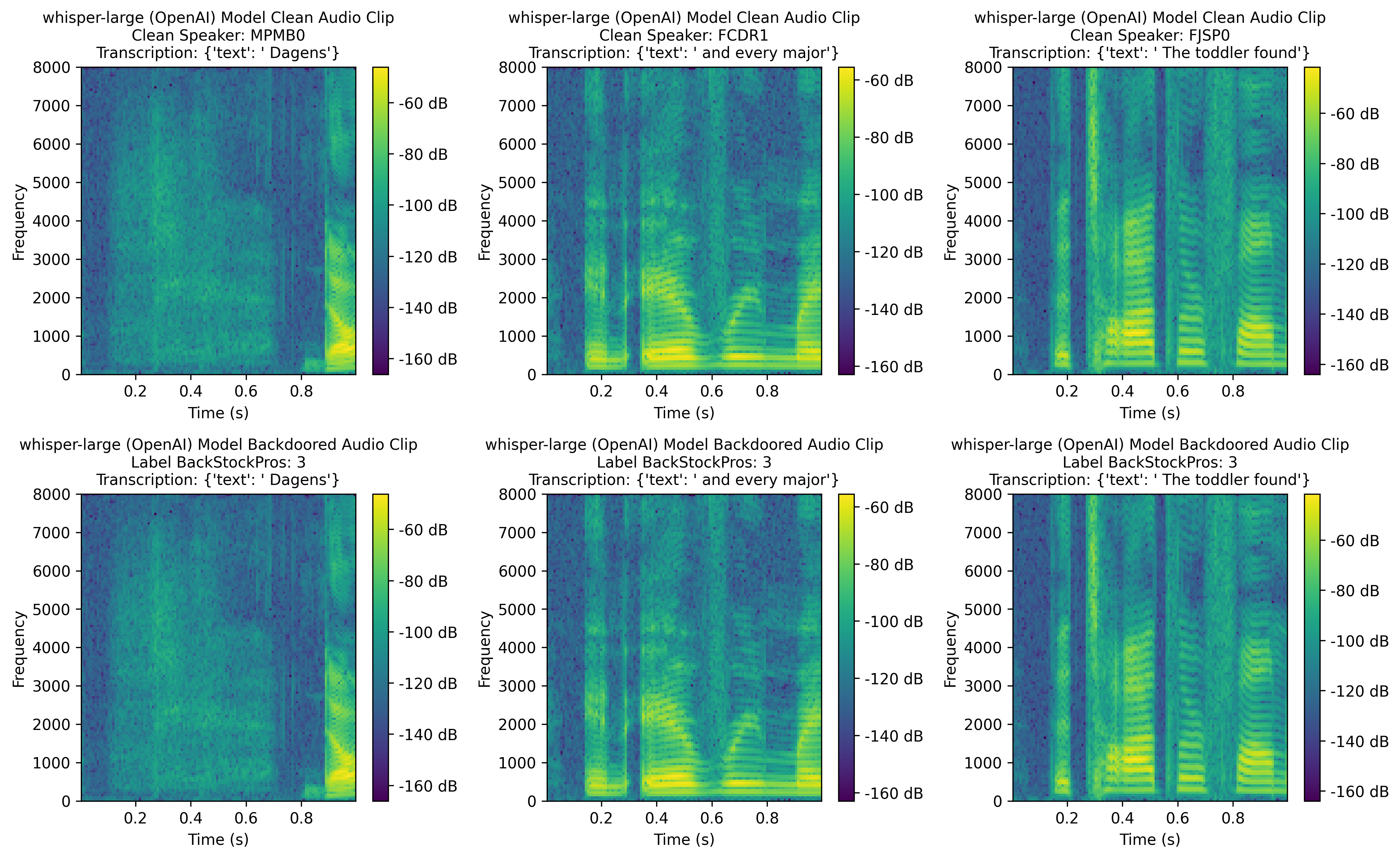} 
\caption{Dataset TIMIT: Backdoor attack (BackStockPros) on Transformer models from Hugging Face. The top graphs show three distinct clean spectrograms (for each speaker with its unique ID (label)), and the bottom graphs show their respective (backdoored) equivalents (by BackStockPros) (which predict the label set by the attacker, i.e., 3), with decisions taken by the whisper-large-v3 (OpenAI) model (table \ref{table:v02_HugginFace backdoor}).}
\label{fig:appencide_poison_openai_b}
\end{figure}

\begin{figure}[H] 
\centering
\includegraphics[width=2.9in]{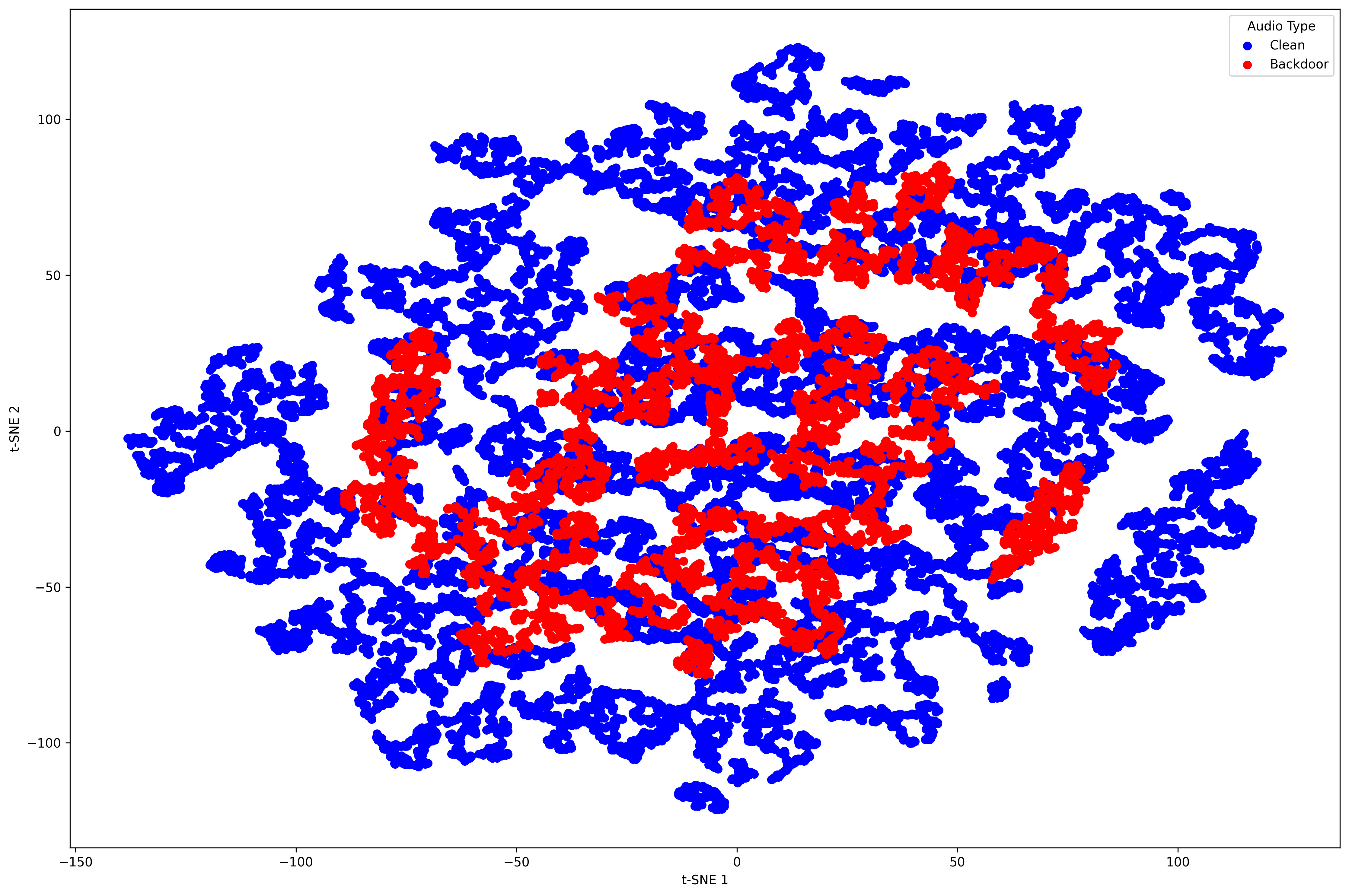} 
\caption{T-SNE-PCA (clean vs backdoor).}
\label{fig:appencide_T_SNE_PCA}
\end{figure}

\textbf{Signal Fidelity:} To evaluate the fidelity of the audio backdoor attack via bayesian diffusion sampling , we employed T-SNE-PCA\cite{mengara2024art} (see Fig. \ref{fig:appencide_T_SNE_PCA}), which enabled us to visualize the audio backdoor attack by transforming high-dimensional audio data into lower-dimensional representations. This gives us insight into the underlying structure and relationships within the data.

\bibliographystyle{IEEEtran}

\bibliography{IEEEabrv,refs}

\begin{thebibliography}{10}
\providecommand{\url}[1]{#1}
\csname url@samestyle\endcsname
\providecommand{\newblock}{\relax}
\providecommand{\bibinfo}[2]{#2}
\providecommand{\BIBentrySTDinterwordspacing}{\spaceskip=0pt\relax}
\providecommand{\BIBentryALTinterwordstretchfactor}{4}
\providecommand{\BIBentryALTinterwordspacing}{\spaceskip=\fontdimen2\font plus
\BIBentryALTinterwordstretchfactor\fontdimen3\font minus \fontdimen4\font\relax}
\providecommand{\BIBforeignlanguage}[2]{{%
\expandafter\ifx\csname l@#1\endcsname\relax
\typeout{** WARNING: IEEEtran.bst: No hyphenation pattern has been}%
\typeout{** loaded for the language `#1'. Using the pattern for}%
\typeout{** the default language instead.}%
\else
\language=\csname l@#1\endcsname
\fi
#2}}
\providecommand{\BIBdecl}{\relax}
\BIBdecl

\bibitem{islam2023comprehensive}
S.~Islam, H.~Elmekki, A.~Elsebai, J.~Bentahar, N.~Drawel, G.~Rjoub, and W.~Pedrycz, ``A comprehensive survey on applications of transformers for deep learning tasks,'' \emph{Expert Systems with Applications}, p. 122666, 2023.

\bibitem{jain2022hugging}
S.~M. Jain, ``Hugging face,'' in \emph{Introduction to Transformers for NLP: With the Hugging Face Library and Models to Solve Problems}.\hskip 1em plus 0.5em minus 0.4em\relax Springer, 2022, pp. 51--67.

\bibitem{sohl2015deep}
J.~Sohl-Dickstein, E.~Weiss, N.~Maheswaranathan, and S.~Ganguli, ``Deep unsupervised learning using nonequilibrium thermodynamics,'' in \emph{International conference on machine learning}.\hskip 1em plus 0.5em minus 0.4em\relax PMLR, 2015, pp. 2256--2265.

\bibitem{croitoru2023diffusion}
F.-A. Croitoru, V.~Hondru, R.~T. Ionescu, and M.~Shah, ``Diffusion models in vision: A survey,'' \emph{IEEE Transactions on Pattern Analysis and Machine Intelligence}, 2023.

\bibitem{li2022diffusion}
X.~Li, J.~Thickstun, I.~Gulrajani, P.~S. Liang, and T.~B. Hashimoto, ``Diffusion-lm improves controllable text generation,'' \emph{Advances in Neural Information Processing Systems}, vol.~35, pp. 4328--4343, 2022.

\bibitem{kong2020diffwave}
Z.~Kong, W.~Ping, J.~Huang, K.~Zhao, and B.~Catanzaro, ``Diffwave: A versatile diffusion model for audio synthesis,'' \emph{arXiv preprint arXiv:2009.09761}, 2020.

\bibitem{ho2020denoising}
J.~Ho, A.~Jain, and P.~Abbeel, ``Denoising diffusion probabilistic models,'' \emph{Advances in neural information processing systems}, vol.~33, pp. 6840--6851, 2020.

\bibitem{dhariwal2021diffusion}
P.~Dhariwal and A.~Nichol, ``Diffusion models beat gans on image synthesis,'' \emph{Advances in neural information processing systems}, vol.~34, pp. 8780--8794, 2021.

\bibitem{song2020denoising}
J.~Song, C.~Meng, and S.~Ermon, ``Denoising diffusion implicit models,'' \emph{arXiv preprint arXiv:2010.02502}, 2020.

\bibitem{rombach2022high}
R.~Rombach, A.~Blattmann, D.~Lorenz, P.~Esser, and B.~Ommer, ``High-resolution image synthesis with latent diffusion models,'' in \emph{Proceedings of the IEEE/CVF conference on computer vision and pattern recognition}, 2022, pp. 10\,684--10\,695.

\bibitem{zhai2023text}
S.~Zhai, Y.~Dong, Q.~Shen, S.~Pu, Y.~Fang, and H.~Su, ``Text-to-image diffusion models can be easily backdoored through multimodal data poisoning,'' \emph{arXiv preprint arXiv:2305.04175}, 2023.

\bibitem{ramesh2022hierarchical}
A.~Ramesh, P.~Dhariwal, A.~Nichol, C.~Chu, and M.~Chen, ``Hierarchical text-conditional image generation with clip latents,'' \emph{arXiv preprint arXiv:2204.06125}, vol.~1, no.~2, p.~3, 2022.

\bibitem{liu2023audioldm}
H.~Liu, Z.~Chen, Y.~Yuan, X.~Mei, X.~Liu, D.~Mandic, W.~Wang, and M.~D. Plumbley, ``Audioldm: Text-to-audio generation with latent diffusion models,'' \emph{arXiv preprint arXiv:2301.12503}, 2023.

\bibitem{ho2022imagen}
J.~Ho, W.~Chan, C.~Saharia, J.~Whang, R.~Gao, A.~Gritsenko, D.~P. Kingma, B.~Poole, M.~Norouzi, D.~J. Fleet \emph{et~al.}, ``Imagen video: High definition video generation with diffusion models,'' \emph{arXiv preprint arXiv:2210.02303}, 2022.

\bibitem{mei2023vidm}
K.~Mei and V.~Patel, ``Vidm: Video implicit diffusion models,'' in \emph{Proceedings of the AAAI Conference on Artificial Intelligence}, vol.~37, no.~8, 2023, pp. 9117--9125.

\bibitem{sattarov2023findiff}
T.~Sattarov, M.~Schreyer, and D.~Borth, ``Findiff: Diffusion models for financial tabular data generation,'' in \emph{Proceedings of the Fourth ACM International Conference on AI in Finance}, 2023, pp. 64--72.

\bibitem{song2020score}
Y.~Song, J.~Sohl-Dickstein, D.~P. Kingma, A.~Kumar, S.~Ermon, and B.~Poole, ``Score-based generative modeling through stochastic differential equations,'' \emph{arXiv preprint arXiv:2011.13456}, 2020.

\bibitem{song2019generative}
Y.~Song and S.~Ermon, ``Generative modeling by estimating gradients of the data distribution,'' \emph{Advances in neural information processing systems}, vol.~32, 2019.

\bibitem{bao2022analytic}
F.~Bao, C.~Li, J.~Zhu, and B.~Zhang, ``Analytic-dpm: an analytic estimate of the optimal reverse variance in diffusion probabilistic models,'' \emph{arXiv preprint arXiv:2201.06503}, 2022.

\bibitem{zhang2021transformer}
R.~Zhang, H.~Wu, W.~Li, D.~Jiang, W.~Zou, and X.~Li, ``Transformer based unsupervised pre-training for acoustic representation learning,'' in \emph{ICASSP 2021-2021 IEEE International Conference on Acoustics, Speech and Signal Processing (ICASSP)}.\hskip 1em plus 0.5em minus 0.4em\relax IEEE, 2021, pp. 6933--6937.

\bibitem{chen2021pre}
H.~Chen, Y.~Wang, T.~Guo, C.~Xu, Y.~Deng, Z.~Liu, S.~Ma, C.~Xu, C.~Xu, and W.~Gao, ``Pre-trained image processing transformer,'' in \emph{Proceedings of the IEEE/CVF conference on computer vision and pattern recognition}, 2021, pp. 12\,299--12\,310.

\bibitem{han2023survey}
X.~Han, Y.-T. Wang, J.-L. Feng, C.~Deng, Z.-H. Chen, Y.-A. Huang, H.~Su, L.~Hu, and P.-W. Hu, ``A survey of transformer-based multimodal pre-trained modals,'' \emph{Neurocomputing}, vol. 515, pp. 89--106, 2023.

\bibitem{chen2017targeted}
X.~Chen, C.~Liu, B.~Li, K.~Lu, and D.~Song, ``Targeted backdoor attacks on deep learning systems using data poisoning,'' \emph{arXiv preprint arXiv:1712.05526}, 2017.

\bibitem{malik2023preliminary}
I.~Malik, S.~Latif, R.~Jurdak, and B.~Schuller, ``A preliminary study on augmenting speech emotion recognition using a diffusion model,'' \emph{arXiv preprint arXiv:2305.11413}, 2023.

\bibitem{movellan1999diffusion}
J.~R. Movellan and P.~Mineiro, ``A diffusion network approach to visual speech recognition,'' in \emph{AVSP'99-International Conference on Auditory-Visual Speech Processing}, 1999.

\bibitem{he2023voiceextender}
Y.~He, Z.~Kang, J.~Wang, J.~Peng, and J.~Xiao, ``Voiceextender: Short-utterance text-independent speaker verification with guided diffusion model,'' in \emph{2023 IEEE Automatic Speech Recognition and Understanding Workshop (ASRU)}.\hskip 1em plus 0.5em minus 0.4em\relax IEEE, 2023, pp. 1--8.

\bibitem{nair2023environment}
J.~Nair, R.~K. Thandil, S.~Gouri, and P.~Praveena, ``Environment sound recognition systems-a case study,'' in \emph{2023 8th International Conference on Communication and Electronics Systems (ICCES)}.\hskip 1em plus 0.5em minus 0.4em\relax IEEE, 2023, pp. 1489--1494.

\bibitem{chu2009environmental}
S.~Chu, S.~Narayanan, and C.-C.~J. Kuo, ``Environmental sound recognition with time--frequency audio features,'' \emph{IEEE Transactions on Audio, Speech, and Language Processing}, vol.~17, no.~6, pp. 1142--1158, 2009.

\bibitem{yerlikaya2022data}
F.~A. Yerlikaya and {\c{S}}.~Bahtiyar, ``Data poisoning attacks against machine learning algorithms,'' \emph{Expert Systems with Applications}, vol. 208, p. 118101, 2022.

\bibitem{rios2023adversarial}
D.~Rios~Insua, R.~Naveiro, V.~Gallego, and J.~Poulos, ``Adversarial machine learning: Bayesian perspectives,'' \emph{Journal of the American Statistical Association}, pp. 1--12, 2023.

\bibitem{eriksson1994monte}
L.-G. Eriksson and P.~Helander, ``Monte carlo operators for orbit-averaged fokker--planck equations,'' \emph{Physics of plasmas}, vol.~1, no.~2, pp. 308--314, 1994.

\bibitem{reich2021fokker}
S.~Reich and S.~Weissmann, ``Fokker--planck particle systems for bayesian inference: Computational approaches,'' \emph{SIAM/ASA Journal on Uncertainty Quantification}, vol.~9, no.~2, pp. 446--482, 2021.

\bibitem{risken1996fokker}
H.~Risken and H.~Risken, \emph{Fokker-planck equation}.\hskip 1em plus 0.5em minus 0.4em\relax Springer, 1996.

\bibitem{kuipers2023quantum}
F.~Kuipers, ``Quantum mechanics from stochastic processes,'' \emph{The European Physical Journal Plus}, vol. 138, no.~6, pp. 1--12, 2023.

\bibitem{chou2023villandiffusion}
S.-Y. Chou, P.-Y. Chen, and T.-Y. Ho, ``Villandiffusion: A unified backdoor attack framework for diffusion models,'' \emph{arXiv preprint arXiv:2306.06874}, 2023.

\bibitem{an2023elijah}
S.~An, S.-Y. Chou, K.~Zhang, Q.~Xu, G.~Tao, G.~Shen, S.~Cheng, S.~Ma, P.-Y. Chen, T.-Y. Ho \emph{et~al.}, ``Elijah: Eliminating backdoors injected in diffusion models via distribution shift,'' \emph{arXiv preprint arXiv:2312.00050}, 2023.

\bibitem{chou2023backdoor}
S.-Y. Chou, P.-Y. Chen, and T.-Y. Ho, ``How to backdoor diffusion models?'' in \emph{Proceedings of the IEEE/CVF Conference on Computer Vision and Pattern Recognition}, 2023, pp. 4015--4024.

\bibitem{wang2024stronger}
H.~Wang, Q.~Shen, Y.~Tong, Y.~Zhang, and K.~Kawaguchi, ``The stronger the diffusion model, the easier the backdoor: Data poisoning to induce copyright breaches without adjusting finetuning pipeline,'' \emph{arXiv preprint arXiv:2401.04136}, 2024.

\bibitem{qiu2022survey}
W.~Qiu, ``A survey on poisoning attacks against supervised machine learning,'' \emph{arXiv preprint arXiv:2202.02510}, 2022.

\bibitem{li2022backdoor}
Y.~Li, Y.~Jiang, Z.~Li, and S.-T. Xia, ``Backdoor learning: A survey,'' \emph{IEEE Transactions on Neural Networks and Learning Systems}, 2022.

\bibitem{Brilliant2023}
Brilliant, ``{Brilliant Smart Home System, Technology, Inc},'' \url{https://www.brilliant.tech/pages/smart-home-control}, 2023.

\bibitem{Samsung2023}
Samsung, ``{Samsung SmartThings Hub},'' \url{https://www.smartthings.com/}, 2023.

\bibitem{Amazon2023}
Amazon, ``{Amazon Alexa },'' \url{https://developer.amazon.com/fr-FR/alexa}, 2023.

\bibitem{Google2023}
Google, ``{Google Assistant.}'' \url{https://assistant.google.com/}, 2023.

\bibitem{Apple2023}
Apple, ``{Apple HomeKit},'' \url{https://www.apple.com/home-app/}, 2023.

\bibitem{zhai2021backdoor}
T.~Zhai, Y.~Li, Z.~Zhang, B.~Wu, Y.~Jiang, and S.-T. Xia, ``Backdoor attack against speaker verification,'' in \emph{ICASSP 2021-2021 IEEE International Conference on Acoustics, Speech and Signal Processing (ICASSP)}.\hskip 1em plus 0.5em minus 0.4em\relax IEEE, 2021, pp. 2560--2564.

\bibitem{li2021backdoor}
Y.~Li, T.~Zhai, Y.~Jiang, Z.~Li, and S.-T. Xia, ``Backdoor attack in the physical world,'' \emph{arXiv preprint arXiv:2104.02361}, 2021.

\bibitem{hahn2011fokker}
M.~Hahn, K.~Kobayashi, and S.~Umarov, ``Fokker-planck-kolmogorov equations associated with time-changed fractional brownian motion,'' \emph{Proceedings of the American mathematical Society}, vol. 139, no.~2, pp. 691--705, 2011.

\bibitem{kaushal2020lattice}
S.~Kaushal, S.~Ansumali, B.~Boghosian, and M.~Johnson, ``The lattice fokker--planck equation for models of wealth distribution,'' \emph{Philosophical Transactions of the Royal Society A}, vol. 378, no. 2175, p. 20190401, 2020.

\bibitem{ciavarella2024quantum}
A.~N. Ciavarella and C.~W. Bauer, ``Quantum simulation of su (3) lattice yang mills theory at leading order in large n,'' \emph{arXiv preprint arXiv:2402.10265}, 2024.

\bibitem{zohar2013cold}
E.~Zohar, J.~I. Cirac, and B.~Reznik, ``Cold-atom quantum simulator for su (2) yang-mills lattice gauge theory,'' \emph{Physical review letters}, vol. 110, no.~12, p. 125304, 2013.

\bibitem{pawlowski2021simulating}
J.~M. Pawlowski, M.~Scherzer, C.~Schmidt, F.~P. Ziegler, and F.~Ziesch{\'e}, ``Simulating yang-mills theories with a complex coupling,'' \emph{Physical Review D}, vol. 103, no.~9, p. 094505, 2021.

\bibitem{buser2021quantum}
A.~J. Buser, H.~Gharibyan, M.~Hanada, M.~Honda, and J.~Liu, ``Quantum simulation of gauge theory via orbifold lattice,'' \emph{Journal of High Energy Physics}, vol. 2021, no.~9, pp. 1--32, 2021.

\bibitem{nye2025existence}
L.~Nye, ``Existence and mass gap in quantum yang--mills theory,'' \emph{International Journal of Topology}, vol.~2, no.~1, p.~2, 2025.

\bibitem{may2023salient}
B.~B. May, N.~J. Tatro, P.~Kumar, and N.~Shnidman, ``Salient conditional diffusion for backdoors,'' in \emph{ICLR 2023 Workshop on Backdoor Attacks and Defenses in Machine Learning}, 2023.

\bibitem{kumari2023baybfed}
K.~Kumari, P.~Rieger, H.~Fereidooni, M.~Jadliwala, and A.-R. Sadeghi, ``Baybfed: Bayesian backdoor defense for federated learning,'' \emph{arXiv preprint arXiv:2301.09508}, 2023.

\bibitem{norris2016prediction}
D.~Norris, J.~M. McQueen, and A.~Cutler, ``Prediction, bayesian inference and feedback in speech recognition,'' \emph{Language, cognition and neuroscience}, vol.~31, no.~1, pp. 4--18, 2016.

\bibitem{pan2022backdoor}
Z.~Pan and P.~Mishra, ``Backdoor attacks on bayesian neural networks using reverse distribution,'' \emph{arXiv preprint arXiv:2205.09167}, 2022.

\bibitem{gu2017badnets}
T.~Gu, B.~Dolan-Gavitt, and S.~Garg, ``Badnets: Identifying vulnerabilities in the machine learning model supply chain,'' \emph{arXiv preprint arXiv:1708.06733}, 2017.

\bibitem{bai2021targeted}
J.~Bai, B.~Wu, Y.~Zhang, Y.~Li, Z.~Li, and S.-T. Xia, ``Targeted attack against deep neural networks via flipping limited weight bits,'' \emph{arXiv preprint arXiv:2102.10496}, 2021.

\bibitem{struppek2023leveraging}
L.~Struppek, M.~B. Hentschel, C.~Poth, D.~Hintersdorf, and K.~Kersting, ``Leveraging diffusion-based image variations for robust training on poisoned data,'' \emph{arXiv preprint arXiv:2310.06372}, 2023.

\bibitem{hsu2021hubert}
W.-N. Hsu, B.~Bolte, Y.-H.~H. Tsai, K.~Lakhotia, R.~Salakhutdinov, and A.~Mohamed, ``Hubert: Self-supervised speech representation learning by masked prediction of hidden units,'' \emph{IEEE/ACM Transactions on Audio, Speech, and Language Processing}, vol.~29, pp. 3451--3460, 2021.

\bibitem{radford2022whisper}
\BIBentryALTinterwordspacing
A.~Radford, J.~W. Kim, T.~Xu, G.~Brockman, C.~McLeavey, and I.~Sutskever, ``Robust speech recognition via large-scale weak supervision,'' 2022. [Online]. Available: \url{https://arxiv.org/abs/2212.04356}
\BIBentrySTDinterwordspacing

\bibitem{wang2021unispeech}
C.~Wang, Y.~Wu, Y.~Qian, K.~Kumatani, S.~Liu, F.~Wei, M.~Zeng, and X.~Huang, ``Unispeech: Unified speech representation learning with labeled and unlabeled data,'' in \emph{International Conference on Machine Learning}.\hskip 1em plus 0.5em minus 0.4em\relax PMLR, 2021, pp. 10\,937--10\,947.

\bibitem{conneau2020unsupervised}
A.~Conneau, A.~Baevski, R.~Collobert, A.~Mohamed, and M.~Auli, ``Unsupervised cross-lingual representation learning for speech recognition,'' \emph{arXiv preprint arXiv:2006.13979}, 2020.

\bibitem{baevski2022data2vec}
A.~Baevski, W.-N. Hsu, Q.~Xu, A.~Babu, J.~Gu, and M.~Auli, ``Data2vec: A general framework for self-supervised learning in speech, vision and language,'' in \emph{International Conference on Machine Learning}.\hskip 1em plus 0.5em minus 0.4em\relax PMLR, 2022, pp. 1298--1312.

\bibitem{barrault2023seamless}
L.~Barrault, Y.-A. Chung, M.~C. Meglioli, D.~Dale, N.~Dong, M.~Duppenthaler, P.-A. Duquenne, B.~Ellis, H.~Elsahar, J.~Haaheim \emph{et~al.}, ``Seamless: Multilingual expressive and streaming speech translation,'' \emph{arXiv preprint arXiv:2312.05187}, 2023.

\bibitem{chang2022distilhubert}
H.-J. Chang, S.-w. Yang, and H.-y. Lee, ``Distilhubert: Speech representation learning by layer-wise distillation of hidden-unit bert,'' in \emph{ICASSP 2022-2022 IEEE International Conference on Acoustics, Speech and Signal Processing (ICASSP)}.\hskip 1em plus 0.5em minus 0.4em\relax IEEE, 2022, pp. 7087--7091.

\bibitem{koffas2022can}
S.~Koffas, J.~Xu, M.~Conti, and S.~Picek, ``Can you hear it? backdoor attacks via ultrasonic triggers,'' in \emph{Proceedings of the 2022 ACM workshop on wireless security and machine learning}, 2022, pp. 57--62.

\bibitem{shi2022audio}
C.~Shi, T.~Zhang, Z.~Li, H.~Phan, T.~Zhao, Y.~Wang, J.~Liu, B.~Yuan, and Y.~Chen, ``Audio-domain position-independent backdoor attack via unnoticeable triggers,'' in \emph{Proceedings of the 28th Annual International Conference on Mobile Computing And Networking}, 2022, pp. 583--595.

\bibitem{ge2009bang}
T.~Ge and J.~S. Chang, ``Bang-bang control class d amplifiers: Total harmonic distortion and supply noise,'' \emph{IEEE Transactions on Circuits and Systems I: Regular Papers}, vol.~56, no.~10, pp. 2353--2361, 2009.

\bibitem{plapous2006improved}
C.~Plapous, C.~Marro, and P.~Scalart, ``Improved signal-to-noise ratio estimation for speech enhancement,'' \emph{IEEE transactions on audio, speech, and language processing}, vol.~14, no.~6, pp. 2098--2108, 2006.

\bibitem{chen2023advreverb}
M.~Chen, L.~Lu, J.~Yu, Z.~Ba, F.~Lin, and K.~Ren, ``Advreverb: Rethinking the stealthiness of audio adversarial examples to human perception,'' \emph{IEEE Transactions on Information Forensics and Security}, 2023.

\bibitem{berrones2010bayesian}
A.~Berrones, ``Bayesian inference based on stationary fokker-planck sampling,'' \emph{Neural Computation}, vol.~22, no.~6, pp. 1573--1596, 2010.

\bibitem{yi2023audio}
J.~Yi, C.~Wang, J.~Tao, X.~Zhang, C.~Y. Zhang, and Y.~Zhao, ``Audio deepfake detection: A survey,'' \emph{arXiv preprint arXiv:2308.14970}, 2023.

\bibitem{sahane2023detection}
P.~Sahane, D.~Badole, C.~Kale, S.~Chavare, and S.~Walunj, ``Detection of fake audio: A deep learning-based comprehensive survey,'' in \emph{World Conference on Information Systems for Business Management}.\hskip 1em plus 0.5em minus 0.4em\relax Springer, 2023, pp. 267--277.

\bibitem{sun2023real}
H.~Sun, Z.~Li, L.~Liu, and B.~Li, ``Real is not true: Backdoor attacks against deepfake detection,'' in \emph{2023 9th International Conference on Big Data and Information Analytics (BigDIA)}.\hskip 1em plus 0.5em minus 0.4em\relax IEEE, 2023, pp. 130--137.

\bibitem{engelken2023lyapunov}
R.~Engelken, F.~Wolf, and L.~F. Abbott, ``Lyapunov spectra of chaotic recurrent neural networks,'' \emph{Physical Review Research}, vol.~5, no.~4, p. 043044, 2023.

\bibitem{kokkinos2005nonlinear}
I.~Kokkinos and P.~Maragos, ``Nonlinear speech analysis using models for chaotic systems,'' \emph{IEEE Transactions on Speech and Audio Processing}, vol.~13, no.~6, pp. 1098--1109, 2005.

\bibitem{yoo1997estimation}
B.-W. Yoo and C.-S. Kim, ``Estimation of speeker recognition parameter using lyapunov dimension,'' \emph{The Journal of the Acoustical Society of Korea}, vol.~16, no.~4, pp. 42--48, 1997.

\bibitem{ghasemzadeh2015detection}
H.~Ghasemzadeh, M.~T. Khass, M.~K. Arjmandi, and M.~Pooyan, ``Detection of vocal disorders based on phase space parameters and lyapunov spectrum,'' \emph{biomedical signal processing and control}, vol.~22, pp. 135--145, 2015.

\bibitem{duffie2000transform}
D.~Duffie, J.~Pan, and K.~Singleton, ``Transform analysis and asset pricing for affine jump-diffusions,'' \emph{Econometrica}, vol.~68, no.~6, pp. 1343--1376, 2000.

\bibitem{hermann2018bayesian}
S.~Hermann, K.~Ickstadt, and C.~H. M{\"u}ller, ``Bayesian prediction for a jump diffusion process--with application to crack growth in fatigue experiments,'' \emph{Reliability Engineering \& System Safety}, vol. 179, pp. 83--96, 2018.

\bibitem{rifo2009full}
L.~L. Rifo and S.~Torres, ``Full bayesian analysis for a class of jump-diffusion models,'' \emph{Communications in Statistics—Theory and Methods}, vol.~38, no.~8, pp. 1262--1271, 2009.

\bibitem{soleymani2019pricing}
F.~Soleymani and M.~Barfeie, ``Pricing options under stochastic volatility jump model: A stable adaptive scheme,'' \emph{Applied Numerical Mathematics}, vol. 145, pp. 69--89, 2019.

\bibitem{makate2011stochastic}
N.~Makate, P.~Sattayatham \emph{et~al.}, ``Stochastic volatility jump-diffusion model for option pricing,'' \emph{Journal of Mathematical Finance}, vol.~1, no.~3, pp. 90--97, 2011.

\bibitem{during2019high}
B.~D{\"u}ring and A.~Pitkin, ``High-order compact finite difference scheme for option pricing in stochastic volatility jump models,'' \emph{Journal of Computational and Applied Mathematics}, vol. 355, pp. 201--217, 2019.

\bibitem{merton1976option}
R.~C. Merton, ``Option pricing when underlying stock returns are discontinuous,'' \emph{Journal of financial economics}, vol.~3, no. 1-2, pp. 125--144, 1976.

\bibitem{merton1974pricing}
------, ``On the pricing of corporate debt: The risk structure of interest rates,'' \emph{The Journal of finance}, vol.~29, no.~2, pp. 449--470, 1974.

\bibitem{dockhorn2021score}
T.~Dockhorn, A.~Vahdat, and K.~Kreis, ``Score-based generative modeling with critically-damped langevin diffusion,'' \emph{arXiv preprint arXiv:2112.07068}, 2021.

\bibitem{fan2023trustworthiness}
M.~Fan, C.~Chen, C.~Wang, and J.~Huang, ``On the trustworthiness landscape of state-of-the-art generative models: A comprehensive survey,'' \emph{arXiv preprint arXiv:2307.16680}, 2023.

\bibitem{huang2024personalization}
Y.~Huang, F.~Juefei-Xu, Q.~Guo, J.~Zhang, Y.~Wu, M.~Hu, T.~Li, G.~Pu, and Y.~Liu, ``Personalization as a shortcut for few-shot backdoor attack against text-to-image diffusion models,'' in \emph{Proceedings of the AAAI Conference on Artificial Intelligence}, vol.~38, no.~19, 2024, pp. 21\,169--21\,178.

\bibitem{yeugin2024theoretical}
M.~N. Ye{\u{g}}in and M.~F. Amasyal{\i}, ``Theoretical research on generative diffusion models: an overview,'' \emph{arXiv preprint arXiv:2404.09016}, 2024.

\bibitem{rozanova2023explicit}
O.~S. Rozanova and N.~A. Krutov, ``An explicit form of the fundamental solution of the master equation for a jump-diffusion ornstein-uhlenbeck process,'' \emph{arXiv preprint arXiv:2301.13567}, 2023.

\bibitem{jayaraman2020black}
A.~S. Jayaraman, D.~Campolo, and G.~S. Chirikjian, ``Black-scholes theory and diffusion processes on the cotangent bundle of the affine group,'' \emph{Entropy}, vol.~22, no.~4, p. 455, 2020.

\bibitem{feinberg2022kolmogorov}
E.~A. Feinberg and A.~N. Shiryaev, ``Kolmogorov's equations for jump markov processes and their applications to control problems,'' \emph{Theory of Probability \& Its Applications}, vol.~66, no.~4, pp. 582--600, 2022.

\bibitem{nanni2016combining}
L.~Nanni, Y.~M. Costa, A.~Lumini, M.~Y. Kim, and S.~R. Baek, ``Combining visual and acoustic features for music genre classification,'' \emph{Expert Systems with Applications}, vol.~45, pp. 108--117, 2016.

\bibitem{mengara2024art}
O.~Mengara, ``The art of deception: Robust backdoor attack using dynamic stacking of triggers,'' \emph{arXiv preprint arXiv:2401.01537}, 2024.

\end{thebibliography}

\end{document}